\documentclass{article}

    \usepackage[preprint]{neurips_2024}

\usepackage[utf8]{inputenc} 
\usepackage[T1]{fontenc}    
\usepackage{hyperref}       
\usepackage{url}            
\usepackage{booktabs}       
\usepackage{amsfonts}       
\usepackage{nicefrac}       
\usepackage{microtype}      
\usepackage{xcolor}         
\usepackage{subfig}
\usepackage{graphicx}
\usepackage{mlmath}
\usepackage{bbm}
\usepackage{amsmath}
\usepackage{enumerate}
\usepackage{xcolor}

\title{A rationale from frequency perspective for grokking in training neural network}

\author{%
    Zhangchen Zhou\textsuperscript{\rm 1,2}, 
    Yaoyu Zhang\textsuperscript{\rm 1,2},
    Zhi-Qin John Xu\textsuperscript{\rm 1,2}\thanks{Corresponding author: xuzhiqin@sjtu.edu.cn} \\
    \textsuperscript{\rm 1} Institute of Natural Sciences, MOE-LSC, Shanghai Jiao Tong University\\
    \textsuperscript{\rm 2}  School of Mathematical Sciences, Shanghai Jiao Tong University
}

\begin{document}

\maketitle

\begin{abstract}
    Grokking is the phenomenon where neural networks~(NNs) initially fit the training data and later generalize to the test data during training. In this paper, we empirically provide a frequency perspective to explain the emergence of this phenomenon in NNs. The core insight is that the networks initially learn the less salient frequency components present in the test data.
    We observe this phenomenon across both synthetic and real datasets, offering a novel viewpoint for elucidating the grokking phenomenon by characterizing it through the lens of frequency dynamics during the training process. Our empirical frequency-based analysis sheds new light on understanding the grokking phenomenon and its underlying mechanisms.
\end{abstract}

\section{Introduction}


Neural networks (NNs) exhibit a remarkable phenomenon where they can effectively generalize the target function despite being over-parameterized \citep{breiman1995reflections,zhang2016understanding}. Conventionally, it has been believed that during the initial training process, the test and training losses remain relatively consistent. However, in recent years, the grokking phenomenon has been observed in \cite{power2022grokking}, indicating that the training loss decreases significantly faster than the test loss (the test loss may even increase)  initially, and then the test loss decreases rapidly after a certain number of training steps. Interpreting such training dynamics is crucial for understanding the generalization capabilities of NNs.


The grokking phenomenon has been empirically observed across a diverse array of problems, including algorithmic datasets \citep{power2022grokking}, MNIST, IMDb movie reviews, QM9 molecules \citep{liu2022towards}, group operations \citep{chughtai2023toy}, polynomial regression \citep{kumar2024grokking}, sparse parity functions \citep{barak2022hidden,bhattamishra2023simplicity}, and XOR cluster data \citep{xu2023benign}. 

In this work, we provide an explanation for the grokking phenomenon from a frequency perspective, which does not need constraints on data dimensionality or explicit regularization in training dynamics. The key insight is that, in the initial stages of training, NNs learn the less salient frequency components present in the test data. Grokking arises due to a misalignment between the preferred frequency in the training dynamics and the dominant frequency in the test data, which is a consequence of insufficient sampling. We use three examples to illustrate this mechanism.

In the first two examples, we consider one-dimensional synthetic data and high-dimensional parity function, where the training data contains spurious low-frequency components due to aliasing effects caused by insufficient sampling. With common (small) initialization, NNs adhere to the frequency principle (F-Principle) that they often learn data from low to high frequencies. Consequently, during the initial stage of training, the networks learn the misaligned low-frequency components, resulting in an increase in test loss while the training loss decreases. In the third example, we consider the MNIST dataset trained by NNs with large initialization, where high frequencies are preferred in contrast to the F-Principle during the training. Since the MNIST dataset is dominated by low frequency, the test accuracy almost does not increase compared with the fast increase of training accuracy at the initial training stage.

Our work highlights that the distribution of the training data and the frequency preference in training dynamics are critical to the grokking phenomenon. The frequency perspective provides a rationale for the underlying mechanism of how these two factors work together to produce grokking phenomenon. 

\section{Related Works}

\paragraph{Grokking} 

The grokking phenomenon was first proposed by  \cite{power2022grokking} on algorithmic datasets in Transformers \citep{vaswani2017attention}, and \cite{liu2022towards} attributed it to an effective theory of representation learning dynamics. \cite{nanda2022progress} and \cite{furuta2024interpreting} interpreted the grokking phenomenon of algorithmic datasets by investigating the Fourier transform of embedding matrices and logits. \cite{liu2022omnigrok} empirically discovered the grokking phenomenon on datasets beyond algorithmic ones, under the condition of large initialization scale with WD, which was theoretically proved by \cite{lyu2023dichotomy} for homogeneous NNs. \cite{thilak2022slingshot} utilized cyclic phase transitions to explain the grokking phenomenon. \cite{varma2023explaining} explained grokking through circuit efficiency and discovered two novel phenomena called ungrokking and semi-grokking. \cite{barak2022hidden} and \cite{bhattamishra2023simplicity} observed the grokking phenomenon in sparse parity functions. \cite{merrill2023tale} investigated the training dynamics of two-layer neural networks on sparse parity functions and demonstrated that grokking results from the competition between dense and sparse subnetworks. \cite{kumar2024grokking} attributed grokking to the transition in training dynamics from the lazy regime to the rich regime without WD, while we do not require this transition. For example, in our first two examples, we only need the F-Principle without regime transition. \cite{xu2023benign} discovered a grokking phenomenon on XOR cluster data and showed that NNs perform like high-dimensional linear classifiers in the initial training stage when the data dimension is larger than the number of training samples, a constraint that is not necessary for our examples.

\paragraph{Frequency Principle}
The implicit frequency bias of NNs that fit the target function from low to high frequency is named as frequency principle~(F-Principle) \citep{xu2019training,xu2019frequency,xu2022overview} or spectral bias \citep{rahaman2019spectral}. The key insight of the F-principle is that the decay rate of a loss function in the frequency domain derives from the regularity of the activation functions \citep{xu2019frequency}. This phenomenon gives rise to a series of theoretical works in the Neural Tangent Kernel \citep{jacot2018neural}~(NTK) regime \citep{luo2022exact,zhang2019explicitizing,zhang2021linear,cao2021towards,yang2019fine,ronen2019convergence,bordelon2020spectrum} and in general settings \citep{luo2019theory}. \cite{ma2020machine} suggests that the gradient flow of NNs obeys the F-Principle from a continuous viewpoint. Also note that if the initialization of network parameters are too large, F-Principle may not hold in the training dynamics of NNs \citep{xu2019frequency, xu2022overview}.

\section{Preliminaries}
\subsection{Notations}
We denote the dataset as $\mathcal{S}=\{(\vx_i,y_i)\}_{i=1}^{n}$. A NN with trainable parameters $\vtheta$ is denoted as $f(\vx;\vtheta)$. In this article, we consider the Mean Square Error~(MSE):
\begin{equation}
    \ell(\mathcal{S}) = \frac{1}{2n}\sum\limits_{i=1}^{n}(f(\vx_i;\vtheta)-y_i)^2.
    \label{eq:mse_loss}
\end{equation}

\subsection{Nonuniform discrete Fourier transform}
We apply nonuniform discrete Fourier transform~(NUDFT) on the dataset $\mathcal{S}$ in a specific direction $\vd$ in the following way:


\begin{equation}
    \mathcal{F}[\mathcal{S}](\vk) = \frac{1}{n}\sum\limits_{i=1}^{n} y_i\exp({-\text{i} k \vx_i \cdot \vd}).
    \label{eq:NUDFT_project}
\end{equation}

where $\text{i}=\sqrt{-1}$, frequency $\vk=k\vd$.


\section{One-dimensional synthetic data}

\paragraph{Dataset} 
The nonuniform training dataset consisting of $n$ points is constructed as follows:
\begin{enumerate} [i)]
    \item  $10000$ evenly-spaced points in the range $[-\pi,\pi]$ are sampled.
    \item  From this set, we select $20$ points that minimize $|\sin(x)-\sin(6x)|$ and $10$ points that minimize $|\sin(3x)-\sin(6x)|$.
    \item  These $30$ points are then combined with $n-30$ uniformly selected points to form the complete training dataset.
\end{enumerate}

For the uniform dataset, the training data consists of $n$ evenly-spaced points, sampled from the range $[-\pi,\pi]$. The test data, on the other hand, comprises $1000$ evenly-spaced points, sampled from the range $[-\pi,\pi]$.

\paragraph{Experiment Settings} The NN architecture employs a fully-connected structure with four hidden layers of widths $200$-$200$-$200$-$100$. The network is in default initialization in Pytorch. The optimizer employed is Adam, with a learning rate of $2\times10^{-6}$. The target function is $\sin 6x$.

\begin{figure}[htb]
    \centering
    \includegraphics[width=\textwidth]{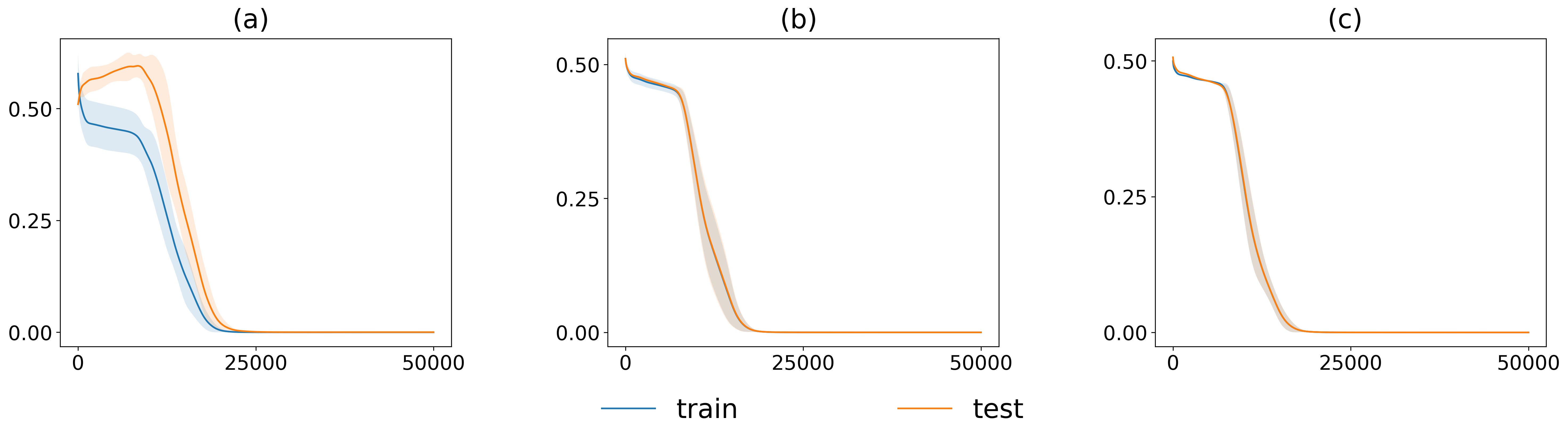}
    \caption{(a) (b) is the train and test loss of $n=65$ and $n=1000$ nonuniform experiment, respectively. (c) is the train and test loss of $n=65$ uniform experiment. The activation function is $\sin x$. Each experiment is averaged over $10$ trials and the shallow parts represent the standard deviation.}
    \label{fig:one_dimension_loss}
\end{figure}

As illustrated in Fig.~\ref{fig:one_dimension_loss} (a), when training on a dataset of $65$ nonuniform points, a distinct grokking phenomenon is observed during the initial stages of the training process, followed by a satisfactory generalization performance in the terminal stage. In contrast, when training on a larger dataset of $1000$ points, as depicted in Fig.~\ref{fig:one_dimension_loss} (b), the grokking phenomenon is notably absent during the initial training phase. This observation aligns with the findings reported in \cite{liu2022omnigrok}, which suggest that larger dataset sizes do not trigger the grokking phenomenon. Furthermore, as shown in Fig.~\ref{fig:one_dimension_loss} (c), we find that when maintaining the data size but uniformly sampling the data, the grokking phenomenon disappears, indicating that the data distribution influences the manifestation of grokking.

\begin{figure}[htb]
    \centering
    \includegraphics[width=\textwidth]{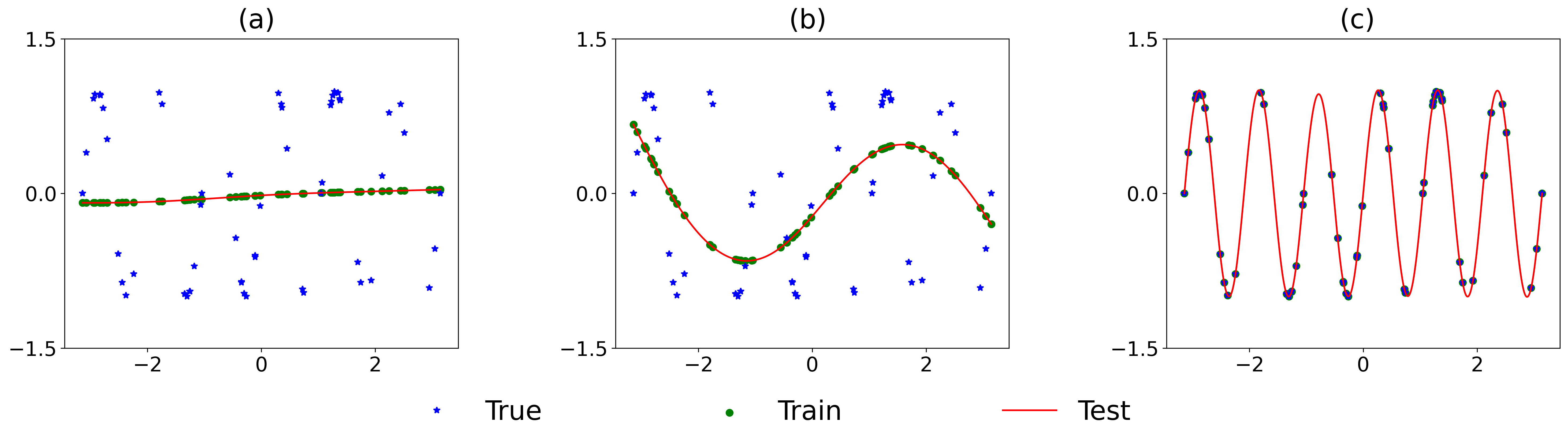}
    \caption{During training with a set of $n=65$ non-uniformly sampled data points, the learned output function evolves across epochs as shown in (a)-(c) for $0$, $2000$, and $35000$ epochs, respectively. The blue stars represent the exact training data points, the green dots are the network's outputs on the training data, and the red curve shows the overall learned output function, drawing on $1000$ evenly-spaced data points.}
    \label{fig:train_test_output_65}
\end{figure}

Upon examining the output of the NN, we observe that with the default initialization, the output is almost zero, as depicted in Fig.~\ref{fig:train_test_output_65} (a). During the training process, the output of the NN resembles $\sin x$, as shown in Fig.~\ref{fig:train_test_output_65} (b). When the loss approaches zero, the network finally fits the target function $\sin 6x$, as illustrated in Fig.~\ref{fig:train_test_output_65} (c). In the subsequent subsection, we will explain this process from the perspective of the frequency domain.

\begin{figure}[htb]
    \centering
    \subfloat{\includegraphics[width=\textwidth]{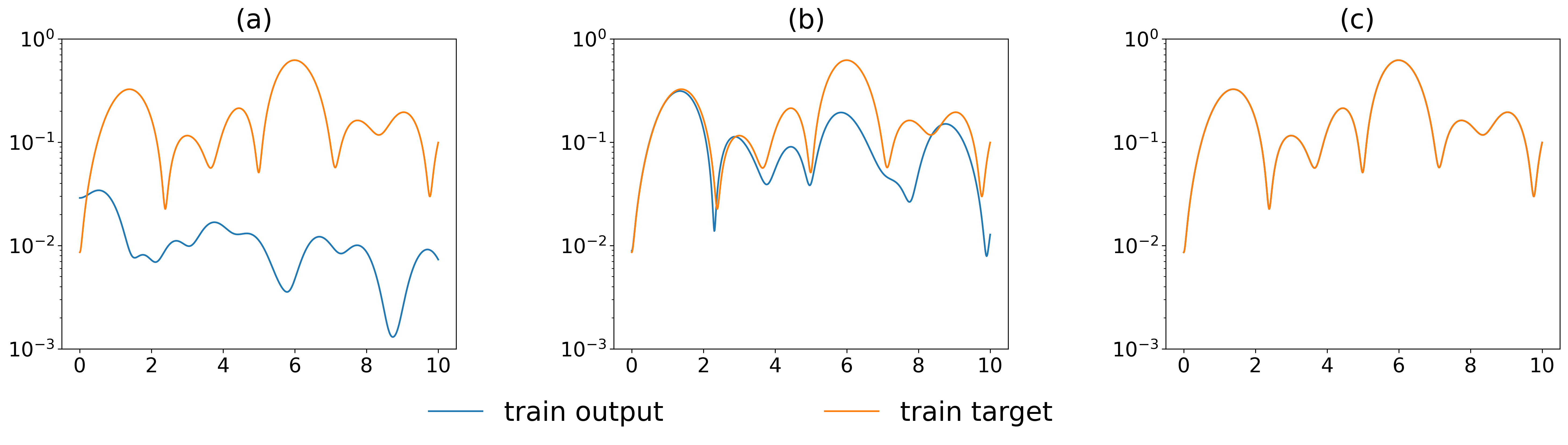}}\\
    \subfloat{\includegraphics[width=\textwidth]{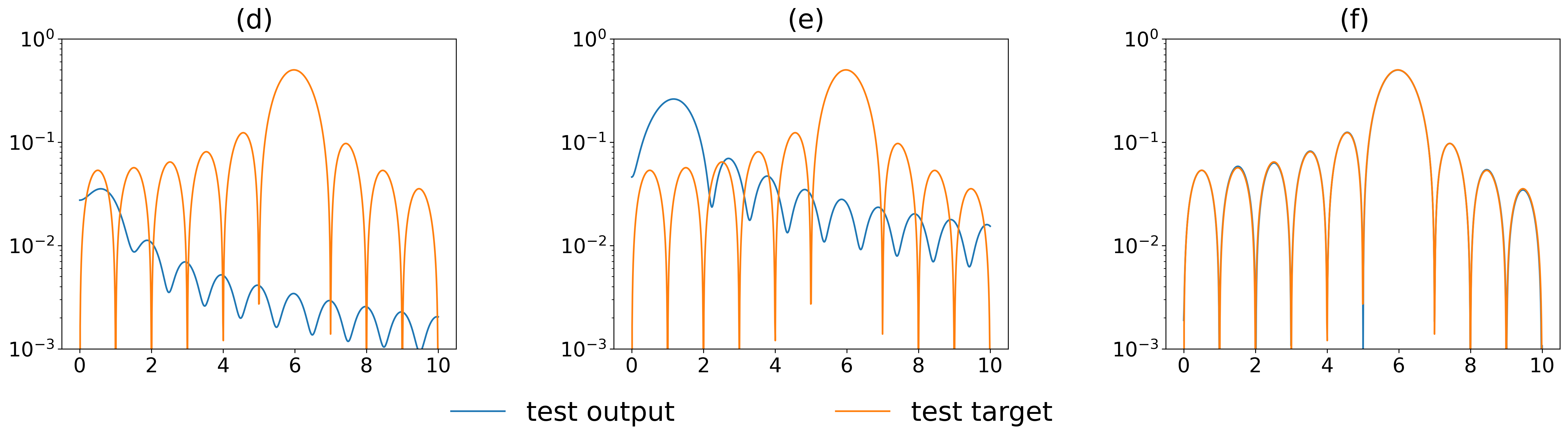}}\\
    \caption{The evolution of the frequency spectrum during training for Fig.~\ref{fig:train_test_output_65}.  The columns from left to right correspond to epochs $0$, $2000$, and $35000$, respectively. The top row illustrates the frequency spectra of the target function (orange solid lines) and the network's output (blue solid lines) on the training data. The bottom row shows the frequency spectra of the target function (orange solid lines) and the network's output (blue solid lines) on the test data. The ordinate represents frequency and the abscissa represents the amplitude of the corresponding frequency components. }
    \label{fig:train_test_frequency_65}
\end{figure}

\subsection{The frequency spectrum of the synthetic data}
We directly utilize Eq.~\eqref{eq:NUDFT_project} to compute the frequency spectrum of the synthetic data and examine the frequency domain. We evenly-spaced sample $1000$ points in the range $[0, 10]$ as the frequency $k$. The peak with the largest amplitude is the genuine frequency $6$. The presence of numerous peaks in the frequency spectrum can be attributed to the fact that the target function multiplies a window function over the interval $[-\pi, \pi]$, which gives rise to spectral leakage.

A key observation is that insufficient and nonuniform sampling leads to a discrepancy between the frequency spectra of the training and test datasets, and the training dataset's spectrum contains a spurious low-frequency component that is not dominant in the test dataset's spectrum. In the case of $n=65$ nonuniform data points, the NUDFT of the training data contrasts starkly with the NUDFT of the true underlying data, as illustrated in Fig.~\ref{fig:train_test_frequency_65} (a) and (d). The training process adheres to the F-Principle, where the NNs initially fit the low-frequency components, as shown in Fig.~\ref{fig:train_test_frequency_65} (b). At this time, as depicted in Fig.~\ref{fig:train_test_frequency_65} (e), the low-frequency components of the output on the test dataset deviate substantially from the exact low-frequency components, inducing an initial ascent in the test loss.

As the training loss approaches zero, we note that the amplitude of the low-frequency components on the test dataset decreases, as illustrated in Fig.~\ref{fig:train_test_frequency_65} (f), while it remains unchanged on the training dataset, as shown in Fig.~\ref{fig:train_test_frequency_65} (c). This phenomenon arises due to two distinct mechanisms for learning the low-frequency components. Initially, following the F-Principle, the network genuinely learns the low-frequency components. Subsequently, as the network learns the high-frequency components, the insufficient sampling induces frequency aliasing onto the low-frequency components, thereby preserving their amplitude on the training dataset.

When training on data with sufficient or uniform sampling, as demonstrated in Fig.~\ref{fig:train_test_frequency_1000} and Fig.~\ref{fig:train_test_frequency_65_uniform} in the appendix, the frequency spectrum of the training dataset aligns with that of the test dataset, thereby mitigating the occurrence of the grokking phenomenon.

Furthermore, this phenomenon is robust across different activation functions, such as $\mathrm{ReLU}$ and $\mathrm{tanh}$, with the results shown in Fig.~\ref{fig:one_dimension_loss_ReLU} and  Fig.~\ref{fig:one_dimension_loss_tanh} in the appendix. The corresponding frequency spectra are depicted in Fig.~\ref{fig:train_test_frequency_65_ReLU}, Fig.~\ref{fig:train_test_frequency_1000_ReLU}, Fig.~\ref{fig:train_test_frequency_65_tanh}, and Fig.~\ref{fig:train_test_frequency_1000_tanh}.

\section{Parity Function}

\paragraph{Parity Function} $\mathcal{I} = \{i_1,i_2,\cdots,i_k\}\subseteq [m]$ is a randomly sampled index set. $(m,\mathcal{I})$ parity function $f_{\mathcal{I}}$ is defined as:

\begin{align}
    f_{\mathcal{I}}: \Omega = \{-1,1\}^{m} & \longrightarrow \{-1,1\}\\
     \vx = \{x_1,x_2,\cdots, x_m \}    & \longmapsto \prod\limits_{i\in \mathcal{I}}x_i
\end{align}

The loss function refers to the MSE loss defined in \eqref{eq:mse_loss}, which directly computes the difference between the labels $y_i$ and the outputs of the NNs $f(\vx_i; \vtheta)$. 
When evaluating the error, we focus on the discrepancy between $y_i$ and $\mathrm{sign}(f(\vx_i; \vtheta))$, effectively treating the problem as a classification task.

\paragraph{Experiment Settings} The experiment targets the $(10, [10])$ parity function, with a total of $2^{10}=1024$ data points. The data is split into training and test datasets with different proportions.
We employ width-$1000$ two-layer fully-connected NN with activation function $\mathrm{ReLU}$. The training is performed using the Adam optimizer with a learning rate of $2\times 10^{-4}$. 

\begin{figure}[htb]
    \centering
    \subfloat[loss]{\includegraphics[width=0.5\textwidth]{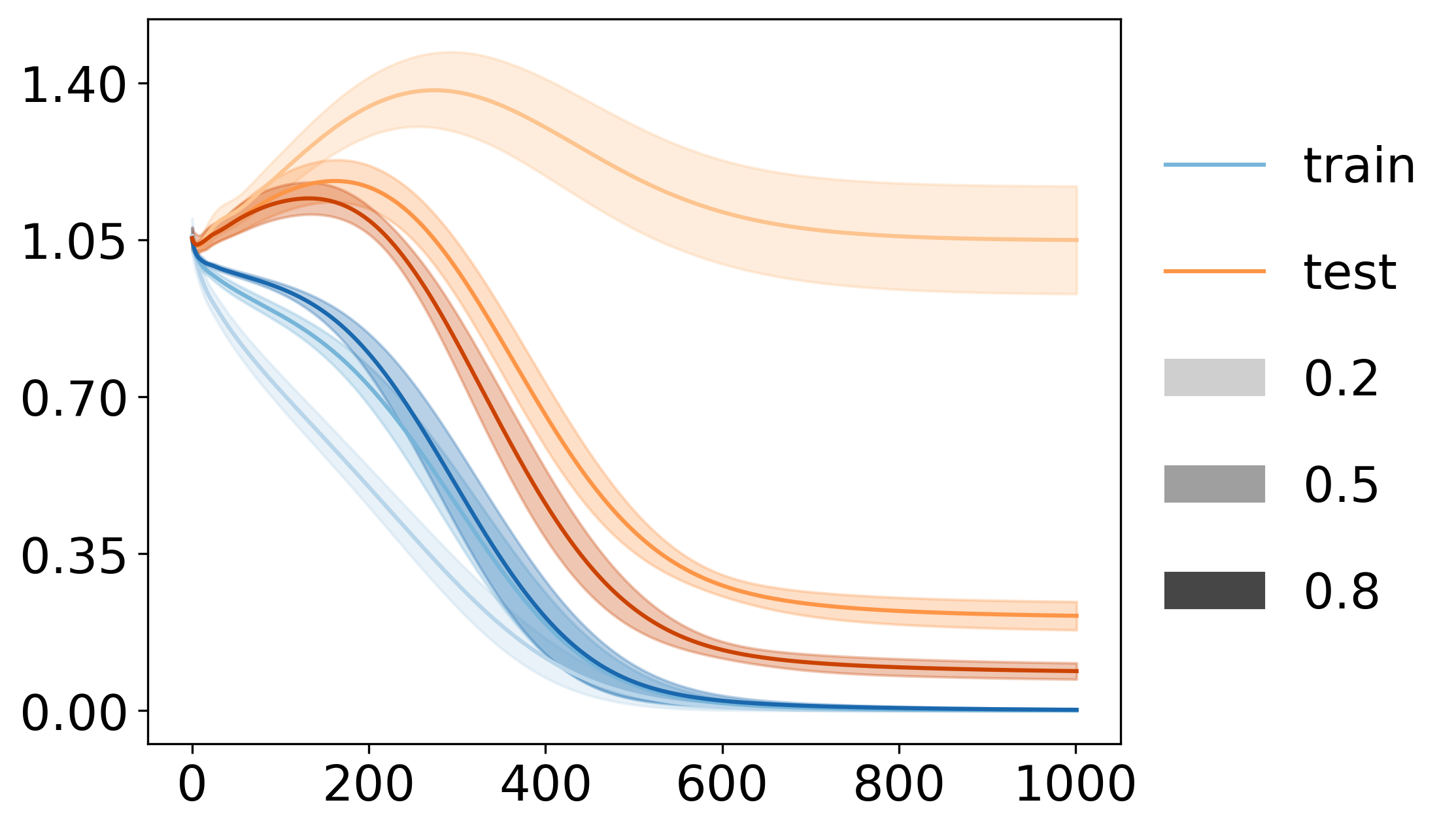}}
    \subfloat[train data frequency]{\includegraphics[width=0.5\textwidth]{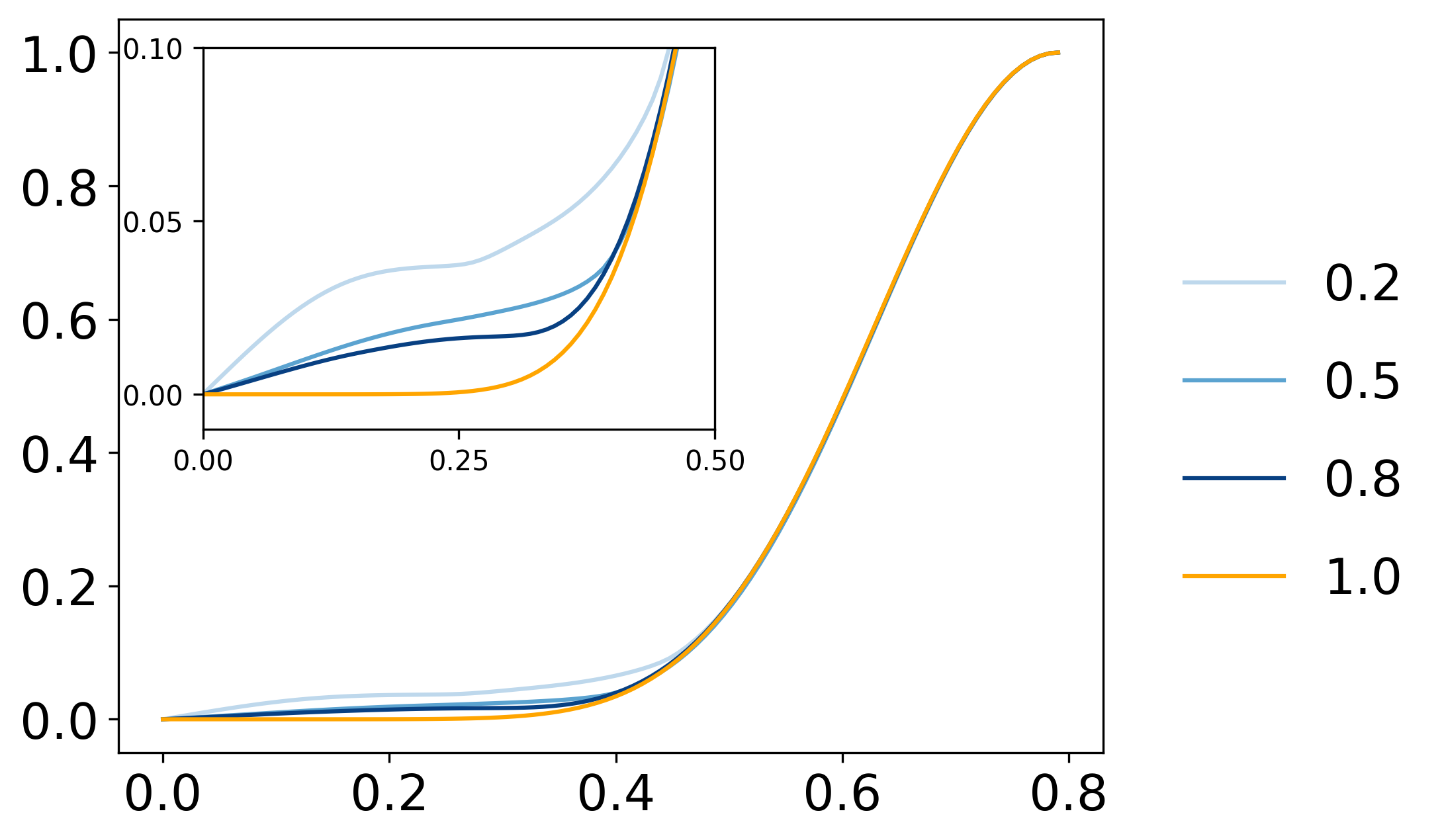}}
    \caption{(a) Loss for ($10,[10]$) parity function with different train data proportion respectively. The blue solid lines are the training dataset, and the orange solid lines are the test dataset.  (b) The frequency spectra of the training set over different proportion ratios. The blue solid lines are the frequency spectra of the training dataset, and the orange solid line is the frequency spectrum of all data. The proportion ratios of the training-test dataset are $0.2$, $0.5$, and $0.8$, from shallow to deep. The ordinate represents frequency and the abscissa represents the amplitude of the corresponding frequency components. Each experiment is averaged over $10$ trials and the shallow parts represent the standard deviation.} 
    \label{fig:parity_loss}
\end{figure}

The grokking phenomenon is observed in the behavior of the loss across different proportions of the training and test set splits, as illustrated in Fig.~\ref{fig:parity_loss} (a). When the training set constitutes a higher proportion of the data, the decline in test loss tends to occur earlier compared to scenarios where the training set has a lower proportion. This observation aligns with the findings reported in \cite{barak2022hidden,bhattamishra2023simplicity}, corroborating the described phenomenon. In the following subsection, we provide an explanation for this phenomenon from a frequency perspective.

\subsection{The frequency spectrum of parity function}
When we directly use Eq.~\eqref{eq:NUDFT_project} to compute the exact frequency spectrum of parity function on $\vxi=(\xi_1,\xi_2,\cdots,\xi_m)$, we obtain

\begin{equation}
    \frac{1}{2^m} \sum\limits_{\vx\in \Omega}\prod\limits_{j=1}^{m}x_j\exp({-\text{i}\vxi}\cdot\vx) = (-\text{i})^{m}\prod\limits_{j=1}^{m}\sin\xi_j.
\end{equation}
To mitigate the computation cost, we only compute on $\vxi=k\times\mathbbm{1}$, where $\mathbbm{1}=(1,1,\cdots,1)$. And $k$ ranges from $0$ to $\frac{\pi}{2}$. 

As illustrated in Fig.~\ref{fig:parity_loss} (b), in the parity function task, when the sampling is insufficient, applying the NUDFT to the training dataset introduces spurious low-frequency components. Moreover, the fewer the sampling points, the more pronounced these low-frequency components become.

\begin{figure}[htb]
    \centering
    \subfloat[loss]{\includegraphics[width=0.5\textwidth]{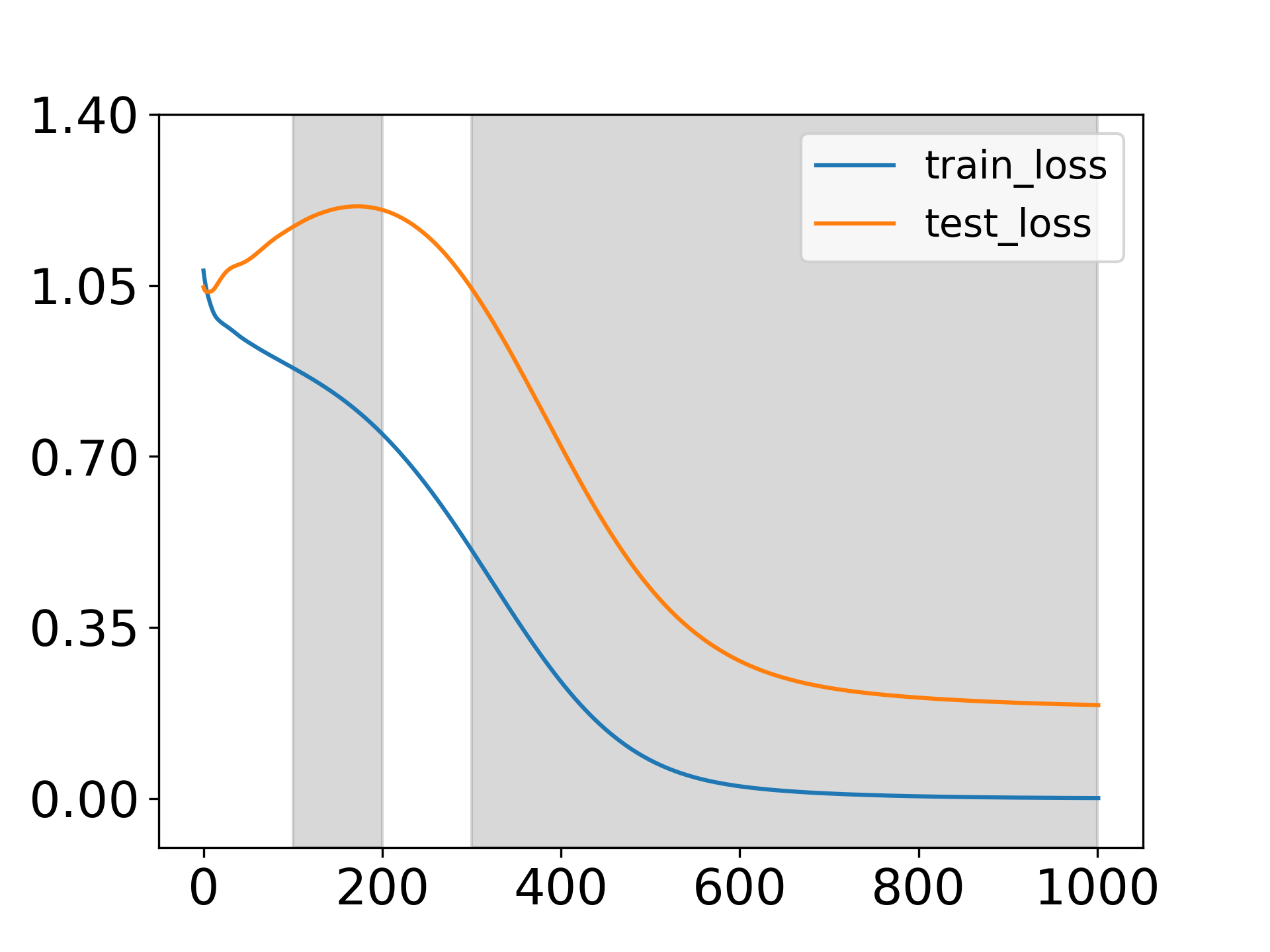}}
    \subfloat[Frequency spectrum]{\includegraphics[width=0.5\textwidth]{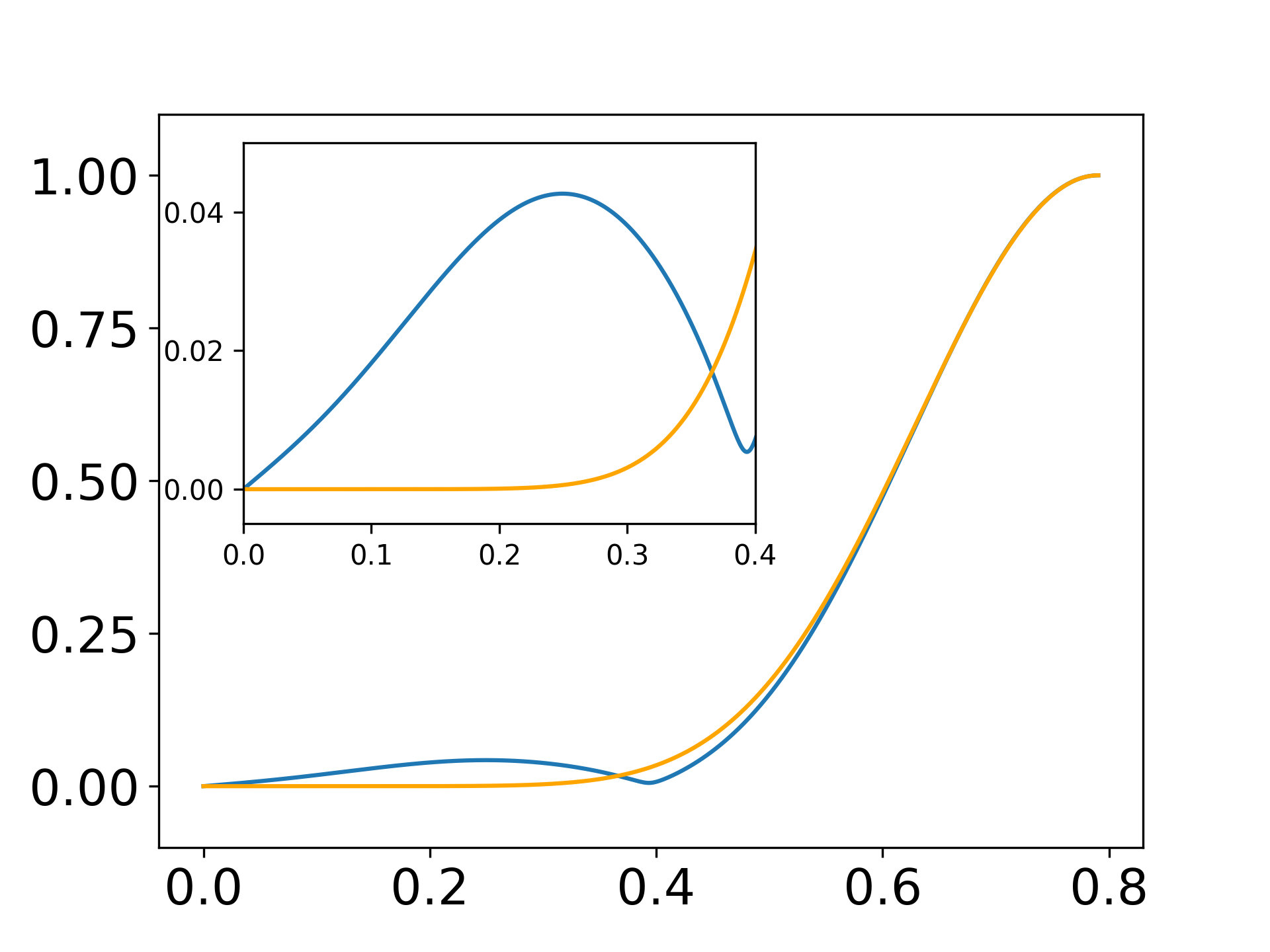}}\\
    
    \subfloat[epoch $100$-$200$]{\includegraphics[width=0.5\textwidth]{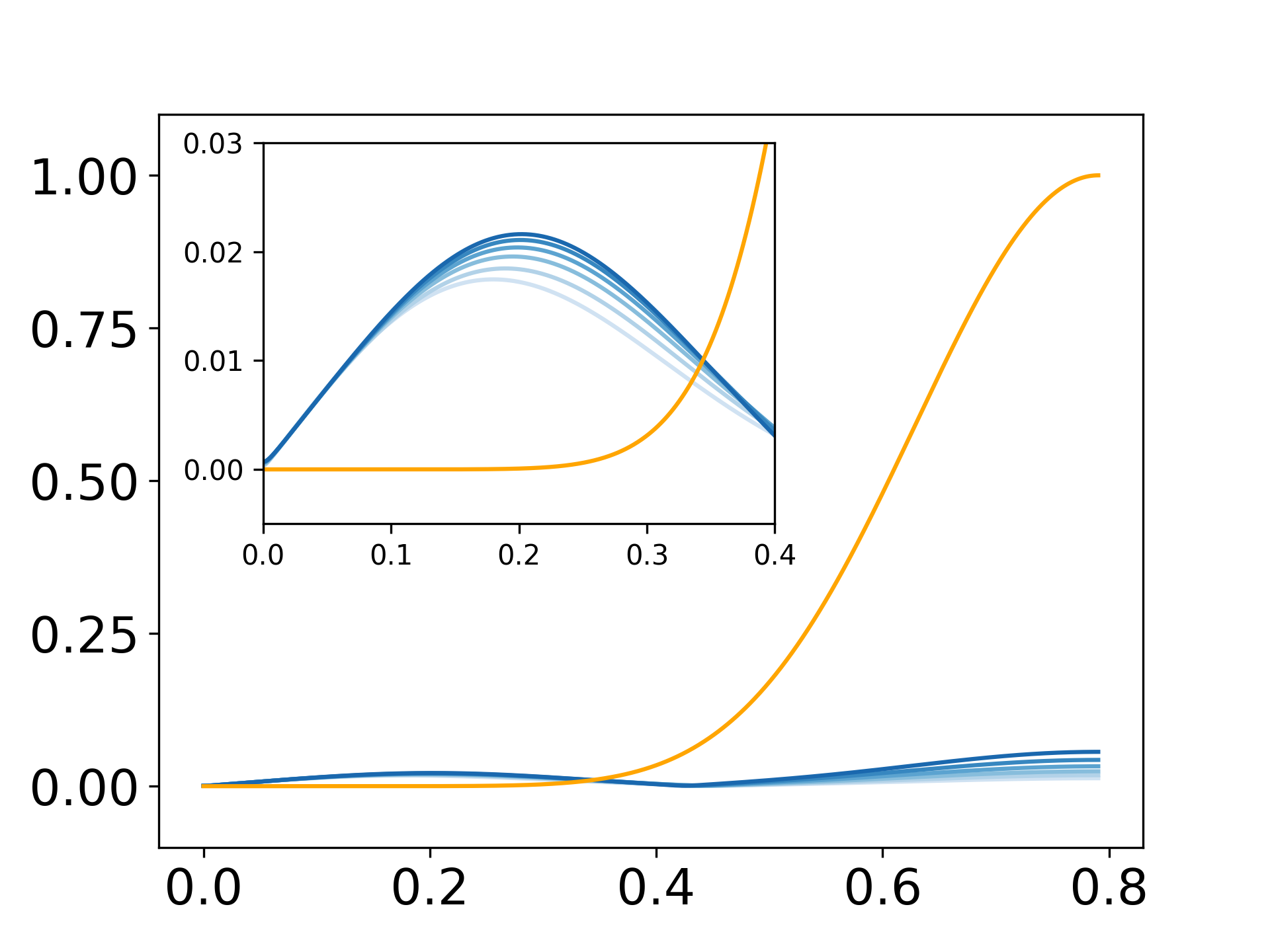}}
    \subfloat[epoch $300$-$1000$]{\includegraphics[width=0.5\textwidth]{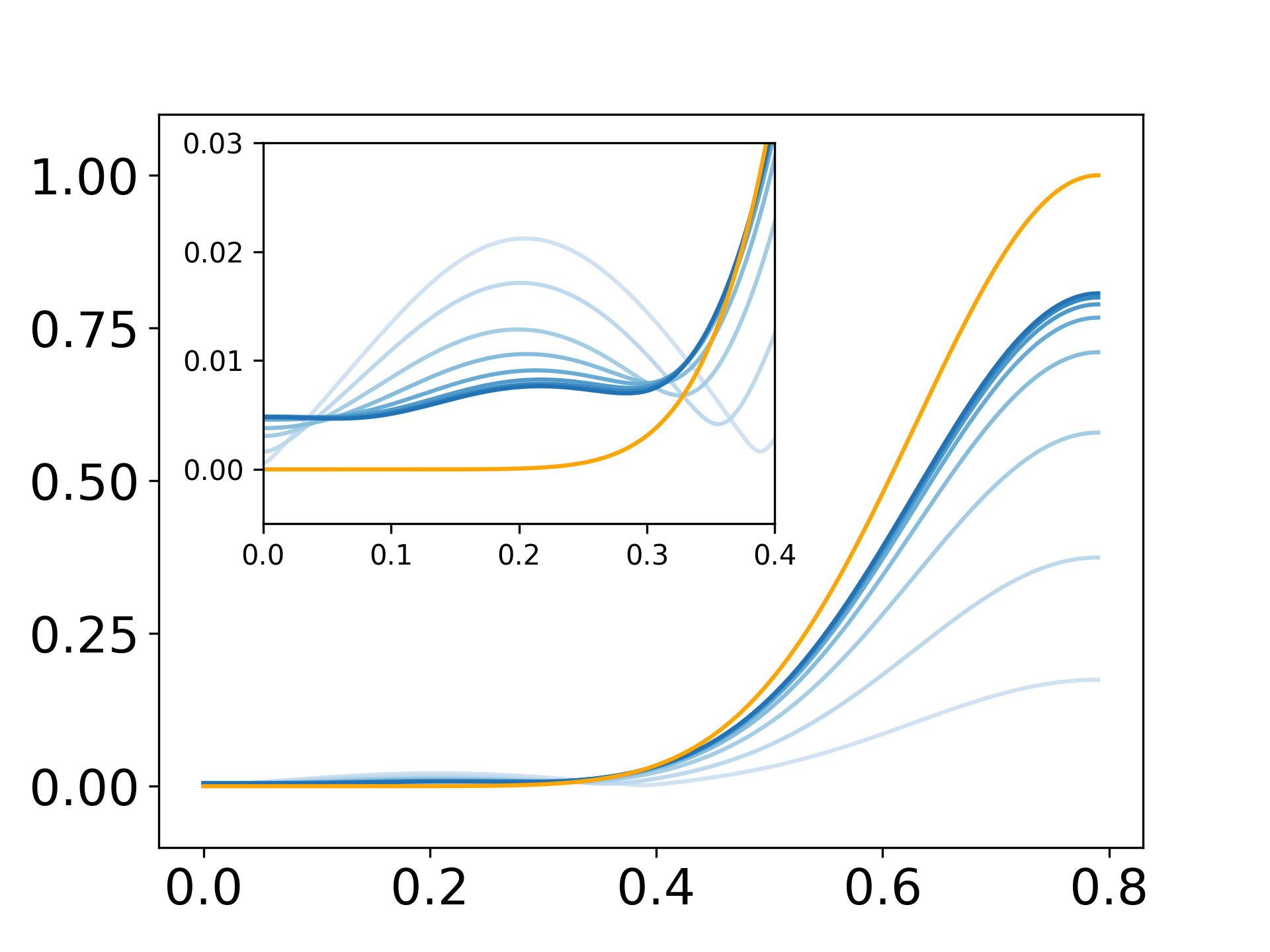}}
    \caption{(a) The train and test loss for a specific experiment with proportion ration $0.5$. The ordinate represents epochs and the abscissa represents the loss. The grey background represents the selected intervals for (c)(d). (b) The frequency spectrum difference between the training dataset and the whole dataset. (c)(d) The frequency spectrum of the whole dataset. The orange solid line is the exact frequency spectrum of the $(10,[10])$ parity function. The blue solid lines are the frequency spectra during the training. From shallow to deep corresponds to an increase in epochs, with (c) recording every $20$ epochs, and (d) recording every $100$ epochs. The ordinate represents frequency and the abscissa represents the amplitude of the corresponding frequency components.}
    \label{fig:parity_process_test}
\end{figure}

To understand the evolution of the frequency domain during the NN's attempt to solve this problem, we illustrate with a concrete example with the proportion $0.5$. Fig.~\ref{fig:parity_process_test} (a) is the loss for this example and Fig.~\ref{fig:parity_process_test} (b) is the frequency spectra of the training dataset and all data. Although the test loss does not approach zero, the test error already goes zero and we regard it as a good generalization as a classification task, as is illustrated in Fig.~\ref{fig:parityfunction_error} (b) in the appendix. During the training process, owing to the F-Principle that governs NNs, the low-frequency components are prioritized and learned first. Consequently, the NN initially fits the spurious low-frequency components arising from insufficient sampling, as is shown in Fig.~\ref{fig:parity_process_test} (c). This results in an initial increase in the test loss, as the low-frequency components on the training dataset are first overfitted. However, as the NN continues to learn the high-frequency components and discovers that utilizing high-frequency components enables a complete fitting of the parity function, we observe a subsequent decline in the low-frequency components, ultimately converging to the true frequency spectrum, which is depicted in Fig.~\ref{fig:parity_process_test} (d).

Based on these findings, we can explain why a lower proportion of training data leads to a later decrease in test loss. When the training data proportion is lower, neural networks spend more time initially fitting the spurious low-frequency components arising from insufficient sampling, before eventually learning the high-frequency components needed to generalize well to the test data.



\section{MNIST with large network initialization}

\paragraph{Experiment Settings} 

We adopt similar experiment settings from \cite{liu2022omnigrok}. The only difference is that we do not utilize explicit weight decay~(WD). $\alpha$ is the scaling factor on the initial parameters. The training set consists of $1000$ points, and the batch size is set to $200$. The NN architecture comprises three fully-connected hidden layers, each with a width of $200$ neurons, employing the ReLU activation function. For large initialization, the Adam optimizer is utilized for training, with a learning rate of $0.001$. For default initialization, the SGD optimizer is employed, with a learning rate of $0.02$.
\begin{figure}[htb]
    \centering
    \subfloat[large initialization $\alpha=8$]{\includegraphics[width=0.5\textwidth]{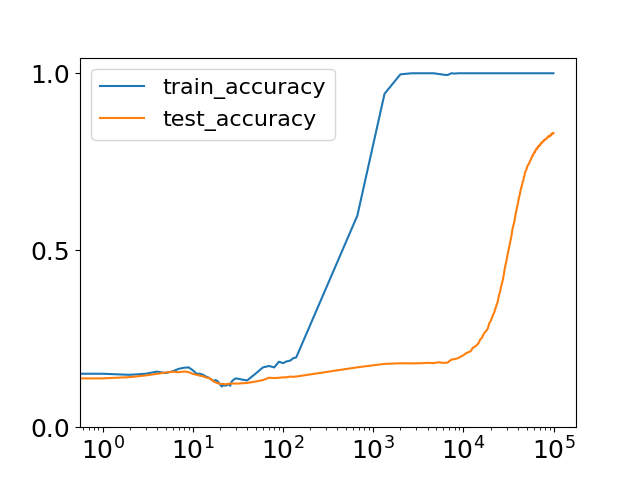}}
    \subfloat[default initialization $\alpha=1$]{\includegraphics[width=0.5\textwidth]{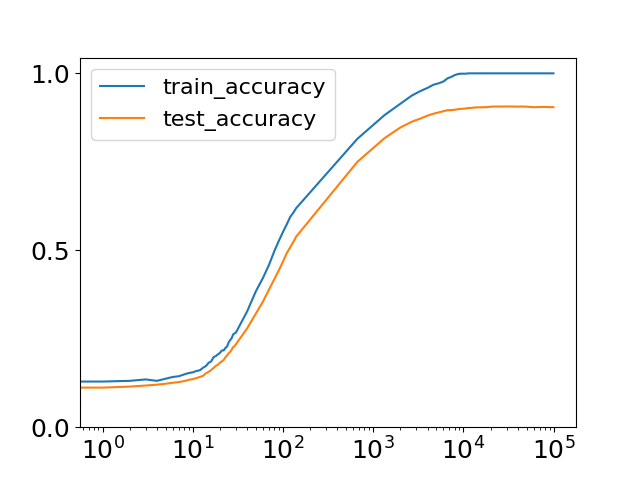}}
    \caption{Accuracy on MNIST dataset for different initialization during the training process. The ordinate represents epochs and the abscissa represents the accuracy.}
    \label{fig:mnist_loss}
\end{figure}

As illustrated in Fig.~\ref{fig:mnist_loss} (a), a grokking phenomenon is observed, suggesting that the explanation provided in \cite{liu2022omnigrok} may not be applicable in this scenario. And when we use default initialization, the grokking phenomenon will not appear, as illustrated in Fig.~\ref{fig:mnist_loss} (b).

\subsection{The frequency spectrum of MNIST} 
In this subsection, we explain the mechanism over the frequency domain.

To mitigate the computation cost, we adopt the projection method introduced in \cite{xu2022overview}. We first compute the principal direction $\vd$ of the dataset (including training and test data) $\mathcal{S}$ through Principal Component Analysis~(PCA) which is intended to preserve more frequency information. Then, we apply Eq.~\eqref{eq:NUDFT_project} to obtain the frequency spectrum of MNIST. The values of $k$ are chosen by sampling $1000$ evenly-spaced points in the interval $[0, 2\pi]$.

\begin{figure}[!ht]
    \centering


    \subfloat{\includegraphics[width=\textwidth]{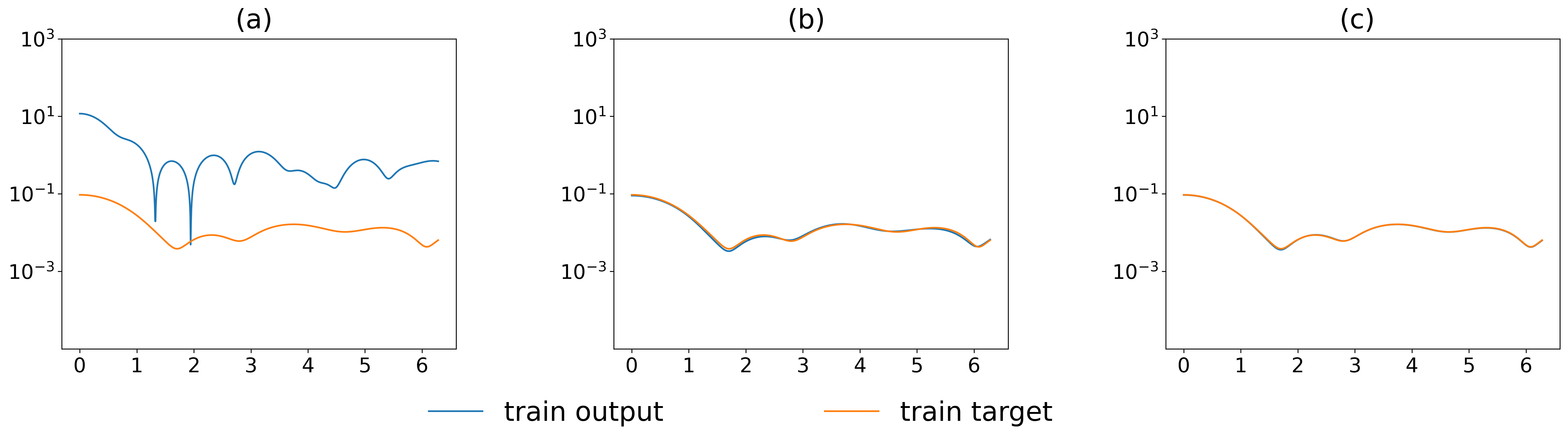}}\\
    \subfloat{\includegraphics[width=\textwidth]{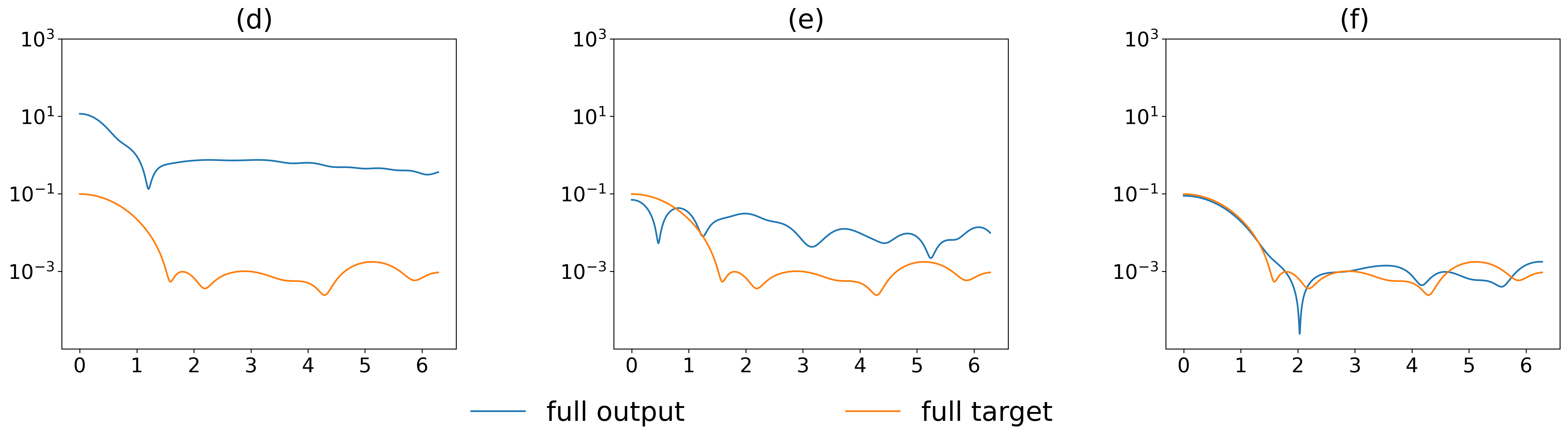}}\\
    \caption{Frequency spectrum evolution during training on the MNIST dataset when $\alpha=8$. The blue solids line represents the frequency spectra of the network's outputs, while the orange solid lines depict the frequency spectra of the target data. The top row shows the frequency spectra on the training set, and the bottom row displays the spectra on the full dataset (train and test combined). The columns from left to right correspond to epochs $0$, $2668$, and $99383$, respectively. The ordinate represents frequency and the abscissa represents the amplitude of the corresponding frequency components.}
    \label{fig:train_test_frequency_mnist}
\end{figure}

\begin{figure}[!ht]
    \centering


    \subfloat{\includegraphics[width=\textwidth]{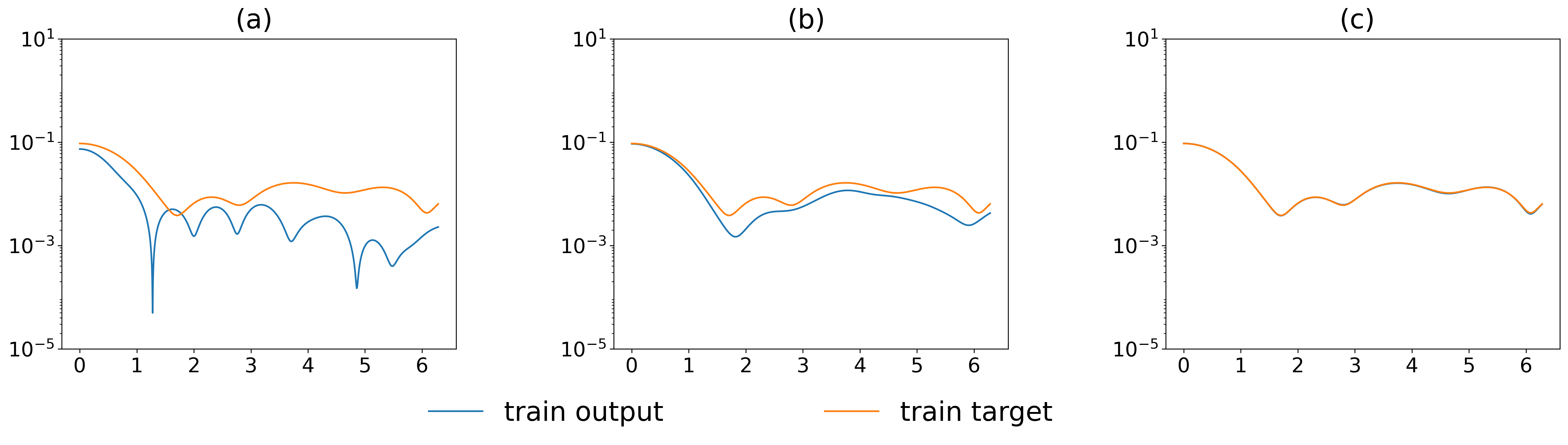}}\\
    \subfloat{\includegraphics[width=\textwidth]{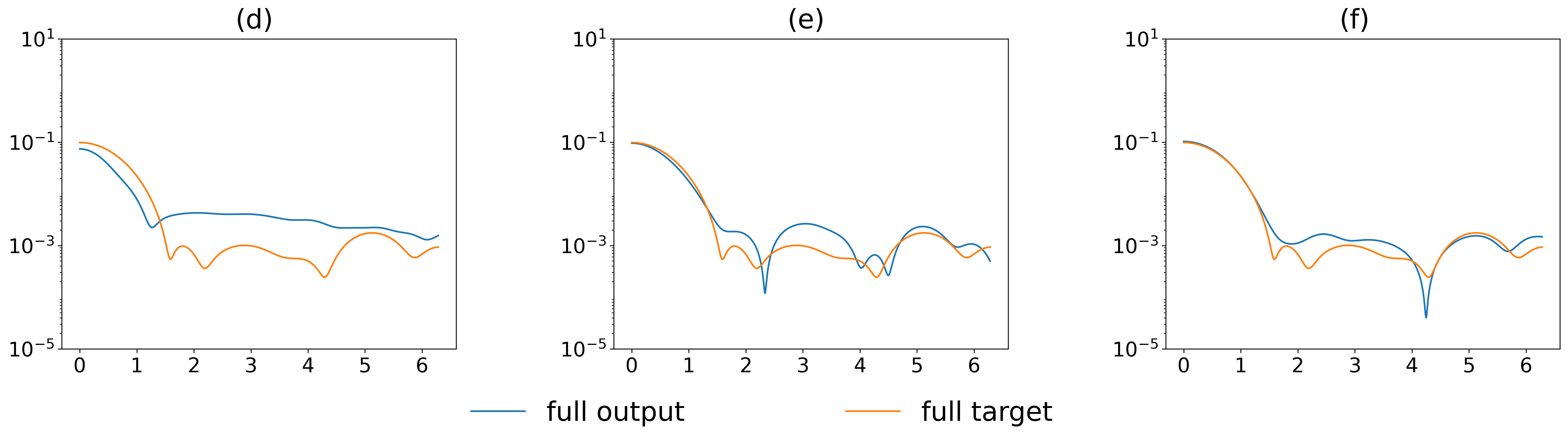}}\\
    \caption{Frequency spectrum evolution during training on the MNIST dataset when $\alpha=1$. The blue solid lines represents the frequency spectra of the network's outputs, while the orange solid lines depict the frequency spectra of the target data. The top row shows the frequency spectra on the training set, and the bottom row displays the spectra on the full dataset (train and test combined). The columns from left to right correspond to epochs $0$, $1334$, and $99383$, respectively. The ordinate represents frequency and the abscissa represents the amplitude of the corresponding frequency components.}
    \label{fig:train_test_frequency_mnist_default}
\end{figure}

In this setting, due to the large initialization adopted, the frequency principle for NNs no longer holds \citep{xu2022overview}, and the low-frequency components will not converge first. During the initial stages of training, the amplitude of the NN's output in the frequency domain substantially exceeds the amplitude of the image itself after the NUDFT, as shown in Fig.~\ref{fig:train_test_frequency_mnist} (a) and (d). After a period of training, the NN fits the frequencies present in the training set, which are primarily caused by high-frequency aliasing due to undersampling (Fig.~\ref{fig:train_test_frequency_mnist} (b)), but it cannot fit the full dataset (Fig.~\ref{fig:train_test_frequency_mnist} (e)). With further training, the NN learns to fit the true high frequencies, at which point the generalization performance of the NN improves, increasing the test set accuracy, as shown in Fig.~\ref{fig:train_test_frequency_mnist} (c) and (f).

However, with default initialization, this phenomenon does not manifest because the NN's fitting pattern adheres to the F-Principle, and image classification functions are dominated by low-frequency components \citep{xu2019frequency}. As shown in Fig.~\ref{fig:train_test_frequency_mnist_default} (a) and (d), the low-frequency spectra of the training and test sets are well-aligned. Consequently, as the network learns the low-frequency components from the training data, it simultaneously captures the low-frequency characteristics of the test set, leading to a simultaneous increase in both training and test accuracy (Fig.~\ref{fig:train_test_frequency_mnist_default} (b) and (e)). Ultimately, the network converges to accurately fit the frequency spectra of both the training set and the complete dataset, as illustrated in Fig.~\ref{fig:train_test_frequency_mnist_default} (c) and (f).

\section{Conclusion and limitations}

In this work, we provided a frequency perspective to elucidate the grokking phenomenon observed in NNs. Our key insight is that during the initial training phase, networks prioritize learning the salient frequency components present in the training data but not dominant in the test data, due to nonuniform and insufficient sampling.

With default initialization, NNs adhere to the F-Principle, fitting the low-frequency components first. Consequently, spurious low frequencies due to insufficient sampling in the training data lead to an initial increase in test loss, as demonstrated in our analyses on one-dimensional synthetic data and parity function learning.  In contrast, with large initialization, NNs fit all frequency components at a similar pace, initially capturing spurious frequencies arising from insufficient sampling, as is shown in the MNIST examples.

This coarse-to-fine processing bears resemblance to the grokking phenomenon observed on XOR cluster data in \cite{xu2023benign}, where high-dimensional linear classifiers were found to first fit the training data. 

However, we only demonstrate this mechanism empirically, leaving a comprehensive theoretical understanding of frequency dynamics during training for future work. Additionally, our mechanism may not apply to the grokking phenomenon observed in language-based data (like the algorithmic datasets).

\section*{Acknowledgments}
Sponsored by the National Key R\&D Program of China  Grant No. 2022YFA1008200, the National Natural Science Foundation of China Grant No. 92270001, 12371511, Shanghai Municipal of Science and Technology Major Project No. 2021SHZDZX0102, and the HPC of School of Mathematical Sciences and the Student Innovation Center, and the Siyuan-1 cluster supported by the Center for High Performance Computing at Shanghai Jiao Tong University.

\bibliography{ref.bib}
\bibliographystyle{elsarticle-num-names}



\newpage 
\appendix

\section{Hardware for the experiments}
\label{app:resource}
All experiments were conducted on a computer equipped with an Intel(R) Xeon(R) Gold 5218 CPU @ 2.30GHz and an NVIDIA GeForce RTX 3090 GPU with 24GB of VRAM. Each experiment can be implemented within several minutes.

\section{One-dimensional synthetic data experiments}

\begin{figure}[!h]
    \centering


    \subfloat{\includegraphics[width=\textwidth]{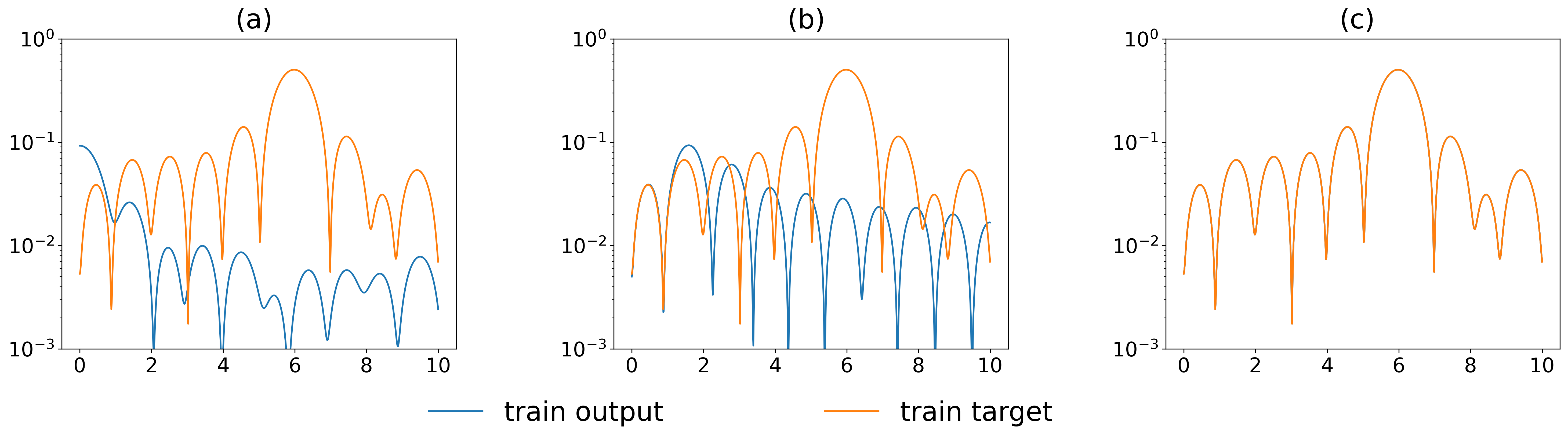}}\\
    \subfloat{\includegraphics[width=\textwidth]{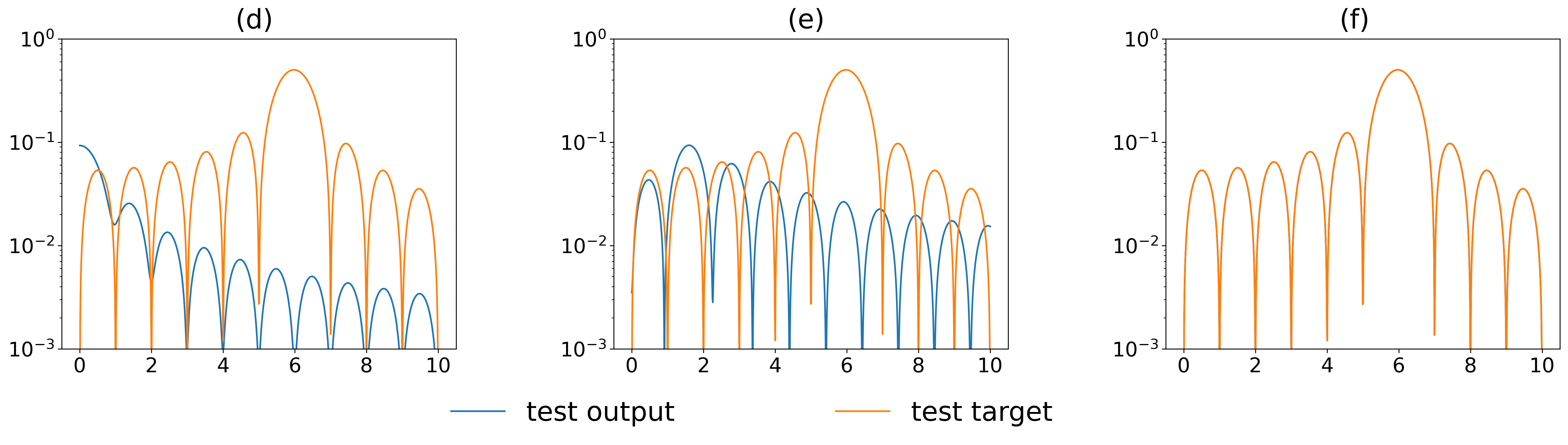}}\\
    \caption{The frequency spectrum of target and output in different epoch during $n=1000$ nonuniform experiment. The columns from left to right correspond to epochs $0$, $2000$, and $35000$, respectively. In the first row, the orange solid line and the blue solid line is about the training target and training output respectively. In the second row, the orange solid line and the blue solid line is about the test target and test output respectively. }
    \label{fig:train_test_frequency_1000}
\end{figure}

\begin{figure}[!h]
    \centering


    \subfloat{\includegraphics[width=\textwidth]{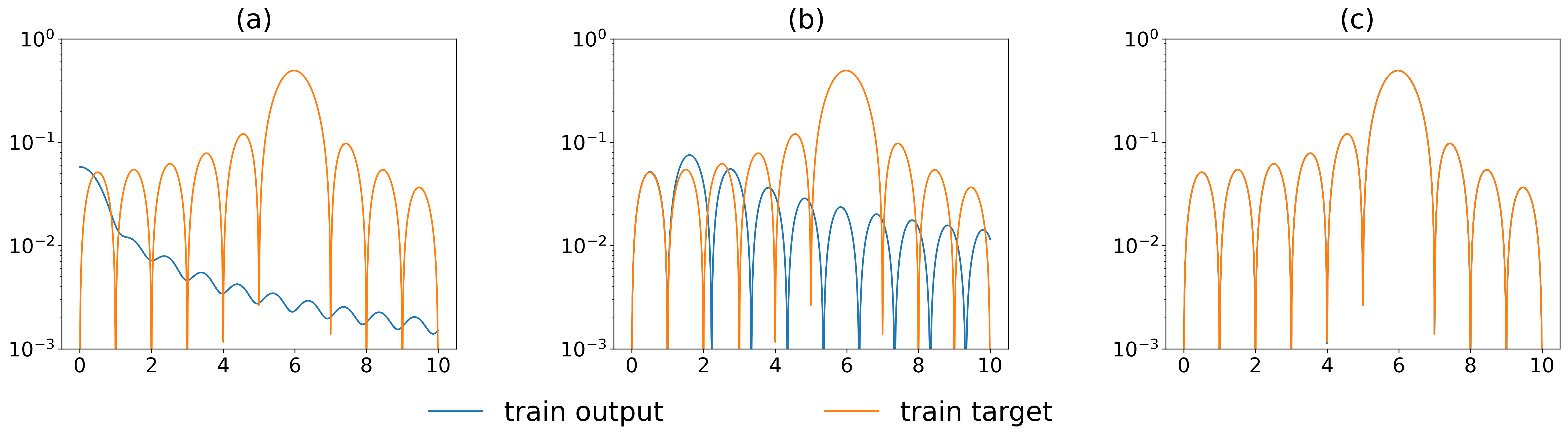}}\\
    \subfloat{\includegraphics[width=\textwidth]{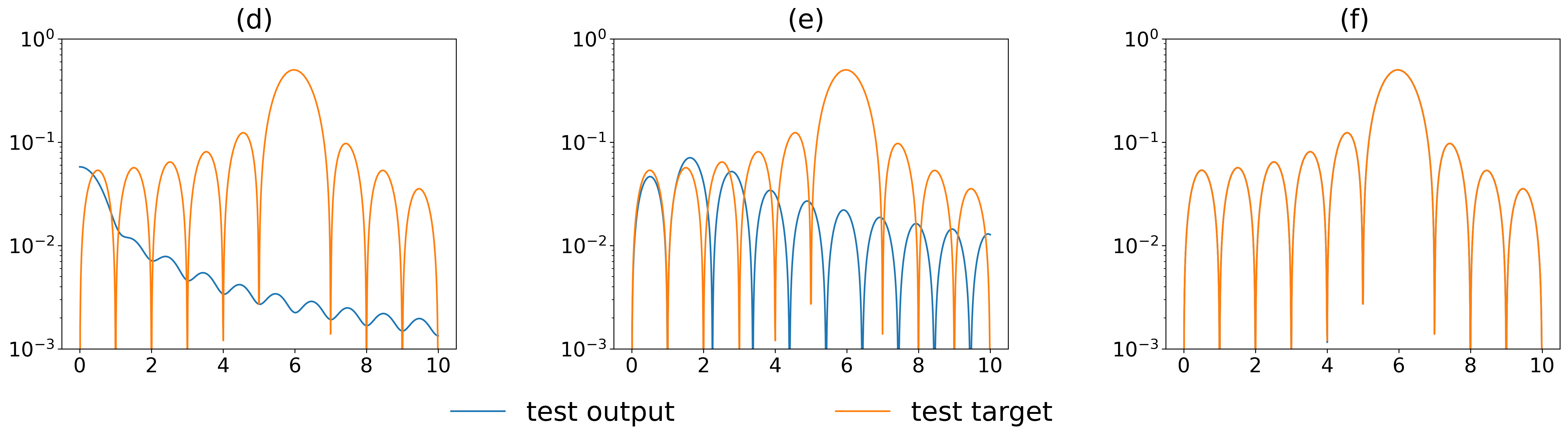}}\\
    \caption{The frequency spectrum of target and output in different epoch during $n=65$ uniform experiment. The columns from left to right correspond to epochs $0$, $2000$, and $35000$, respectively. In the first row, the orange solid line and the blue solid line is about the training target and training output respectively. In the second row, the orange solid line and the blue solid line is about the test target and test output respectively. }
    \label{fig:train_test_frequency_65_uniform}
\end{figure}

Fig.~\ref{fig:train_test_frequency_1000} shows the frequency spectrum of $n=1000$ nonuniform experiment and Fig.~\ref{fig:train_test_frequency_65_uniform} shows the frequency spectrum of $n=65$ uniform experiment with the activation function $\sin x$.

\paragraph{$\mathrm{ReLU}$ activation} The data is generated as in the experiment in the main text with activation function $\mathrm{ReLU} $. The NN architecture employs a fully-connected structure with four layers of widths $200$-$200$-$200$-$100$. The network is in default initialization in Pytorch. The optimizer employed is Adam, with a learning rate of $2\times10^{-5}$.

\paragraph{$\mathrm{Tanh}$ activation} The data is generated as in the experiment in the main text with activation function $\mathrm{Tanh} $. The NN architecture employs a fully-connected structure with four layers of widths $200$-$200$-$200$-$100$. The network is in default initialization in Pytorch. The optimizer employed is Adam, with a learning rate of $2\times10^{-5}$.

\begin{figure}[!h]
    \centering
    \subfloat[$65$]{\includegraphics[width=0.5\textwidth]{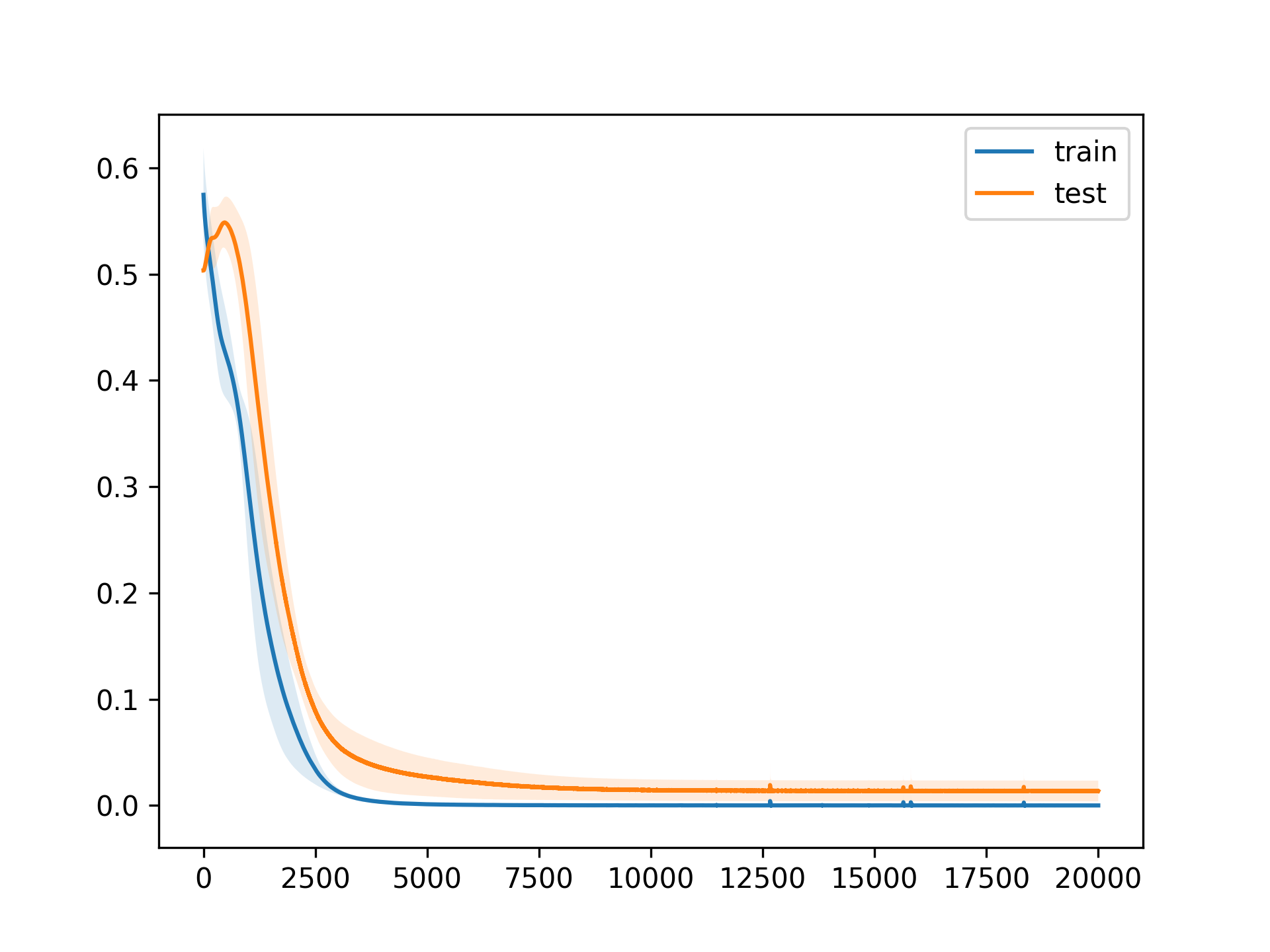}}
    \subfloat[$1000$]{\includegraphics[width=0.5\textwidth]{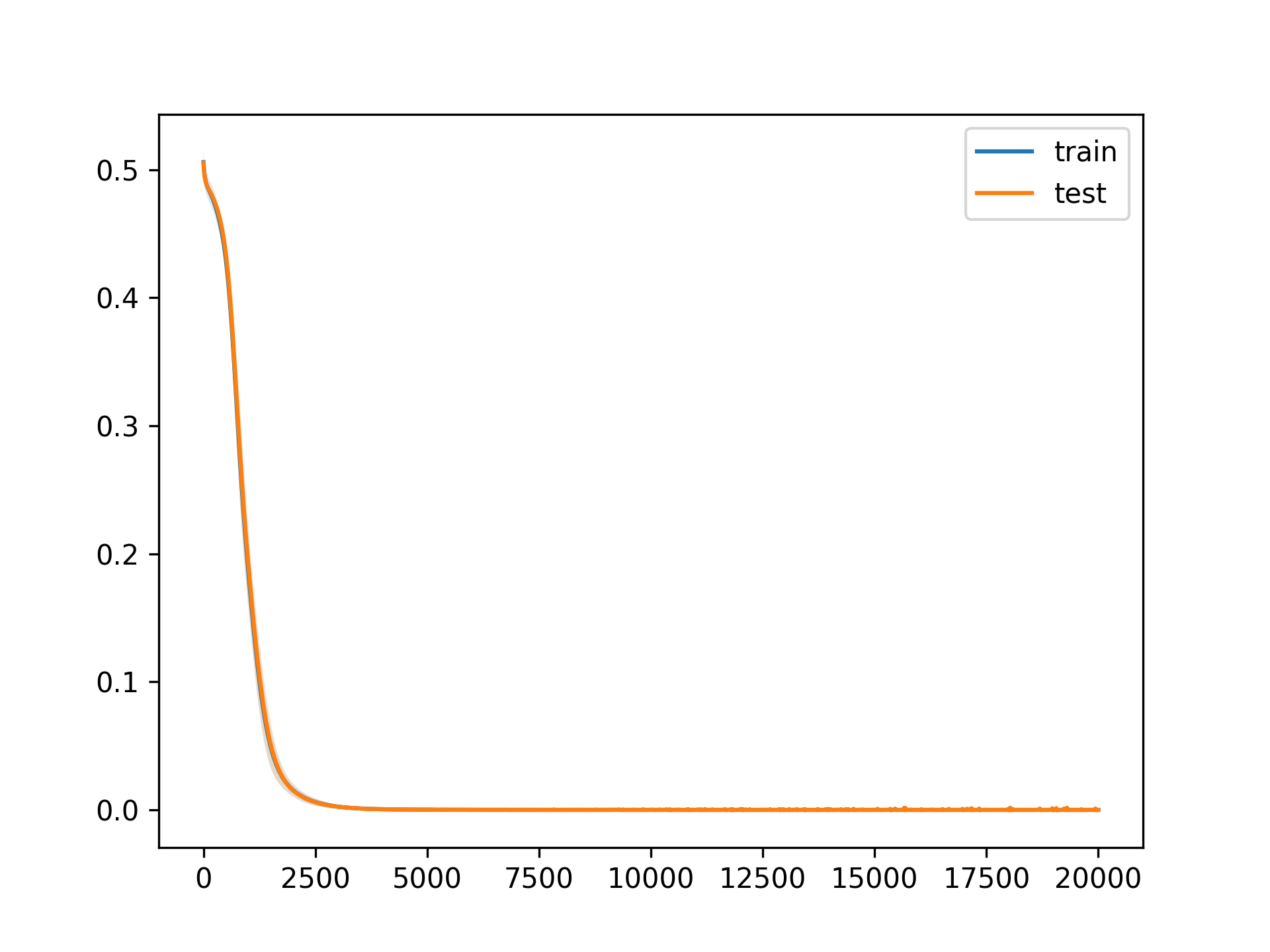}}
    \caption{(a) (b) is the train and test loss of $n=65$ and $n=1000$ nonuniform experiment, respectively. The activation function is $\mathrm{ReLU}$.}
    \label{fig:one_dimension_loss_ReLU}
\end{figure}

\begin{figure}[!h]
    \centering


    \subfloat{\includegraphics[width=\textwidth]{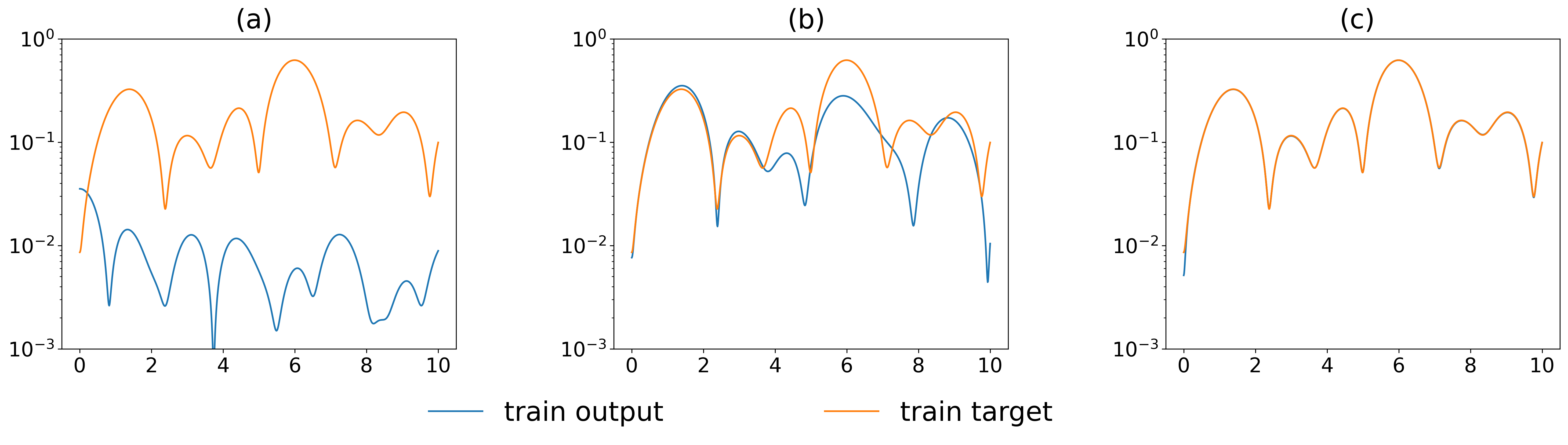}}\\
    \subfloat{\includegraphics[width=\textwidth]{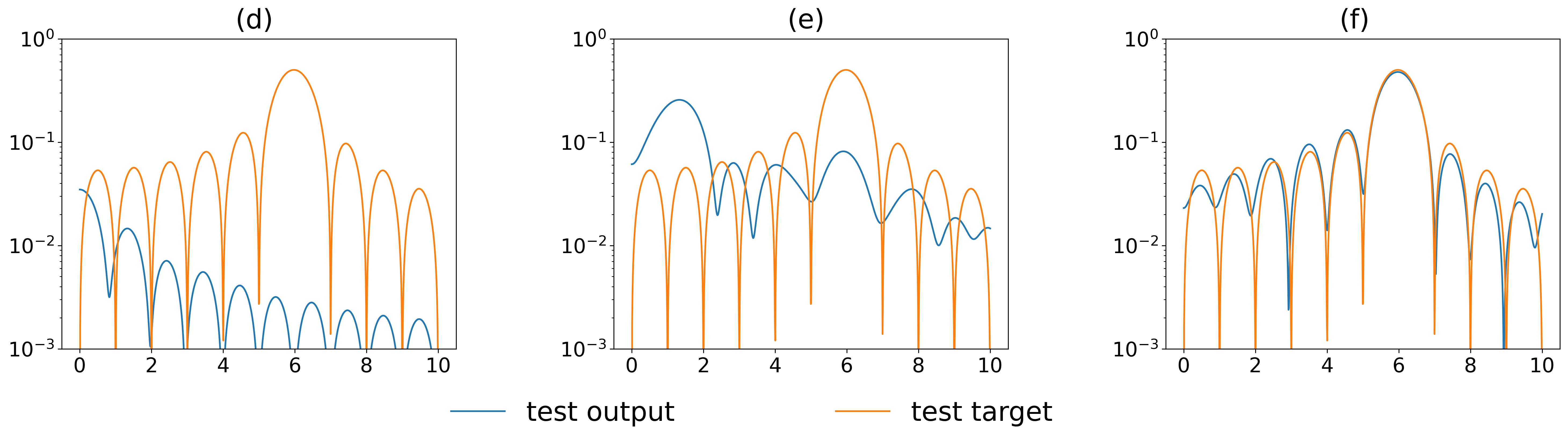}}\\
    \caption{The frequency spectrum during training of Fig.~\ref{fig:one_dimension_loss_ReLU} (a). The columns from left to right correspond to epochs $0$, $1000$, and $19900$, respectively.  The first row illustrates the frequency spectrum of the target function (orange solid line) and the network's output (blue solid line) on the training data. The second row shows the frequency spectrum of the target function (orange solid line) and the network's output (blue solid line) on the test data.}
    \label{fig:train_test_frequency_65_ReLU}
\end{figure}

\begin{figure}[!h]
    \centering


    \subfloat{\includegraphics[width=\textwidth]{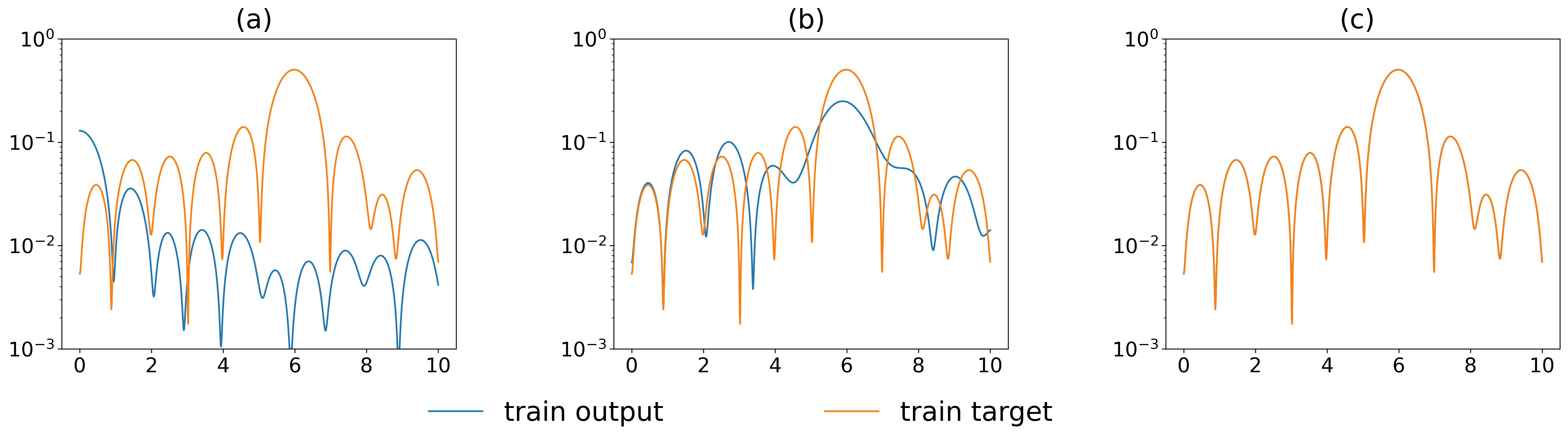}}\\
    \subfloat{\includegraphics[width=\textwidth]{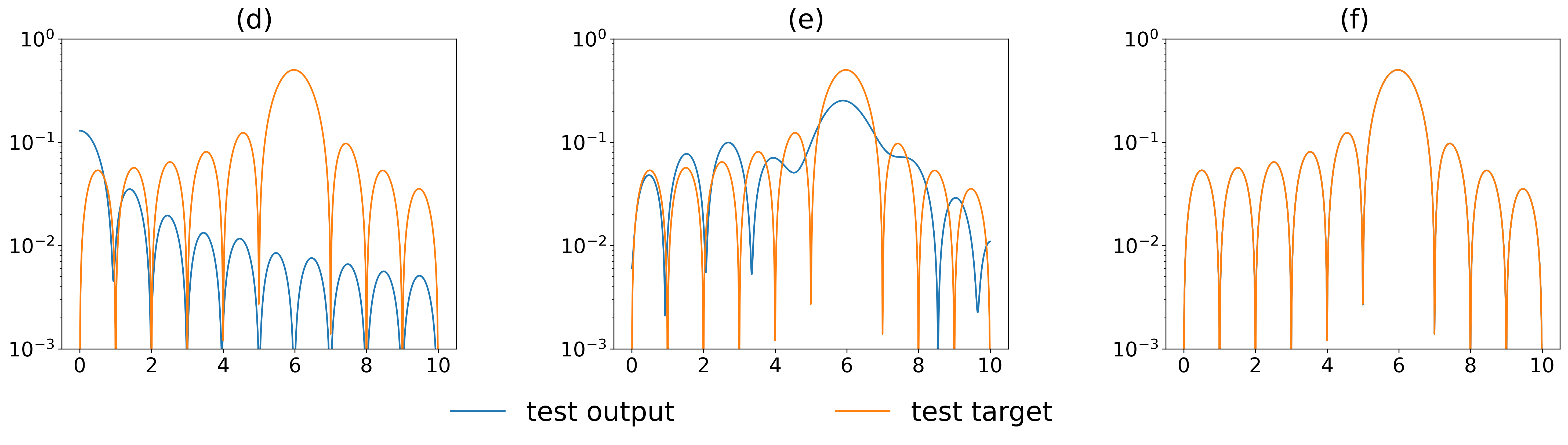}}\\
    \caption{The frequency spectrum during training of Fig.~\ref{fig:one_dimension_loss_ReLU} (b). The columns from left to right correspond to epochs $0$, $1000$, and $19900$, respectively. The first row illustrates the frequency spectrum of the target function (orange solid line) and the network's output (blue solid line) on the training data. The second row shows the frequency spectrum of the target function (orange solid line) and the network's output (blue solid line) on the test data.}
    \label{fig:train_test_frequency_1000_ReLU}
\end{figure}

Fig.~\ref{fig:one_dimension_loss_ReLU} is the loss for $n=65$ and $n=1000$ nonuniform experiments when the activation function is $\mathrm{ReLU} $. Fig.\ref{fig:train_test_frequency_65_ReLU} is the frequency spectrum of $n=65$ nonuniform experiment, and Fig.\ref{fig:train_test_frequency_1000_ReLU} is the frequency spectrum of $n=1000$ nonuniform experiment.

\begin{figure}[!h]
    \centering
    \subfloat[$65$]{\includegraphics[width=0.5\textwidth]{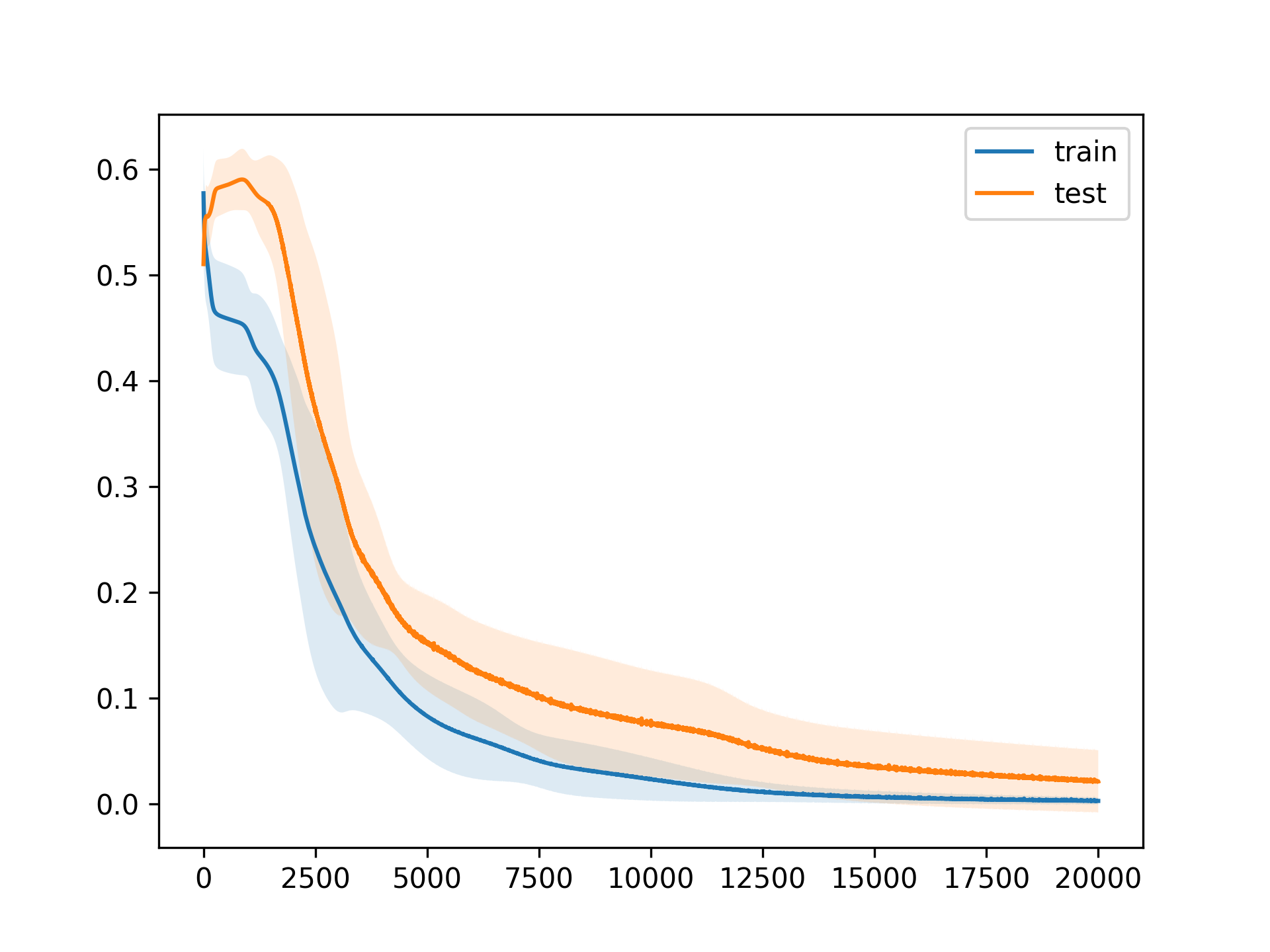}}
    \subfloat[$1000$]{\includegraphics[width=0.5\textwidth]{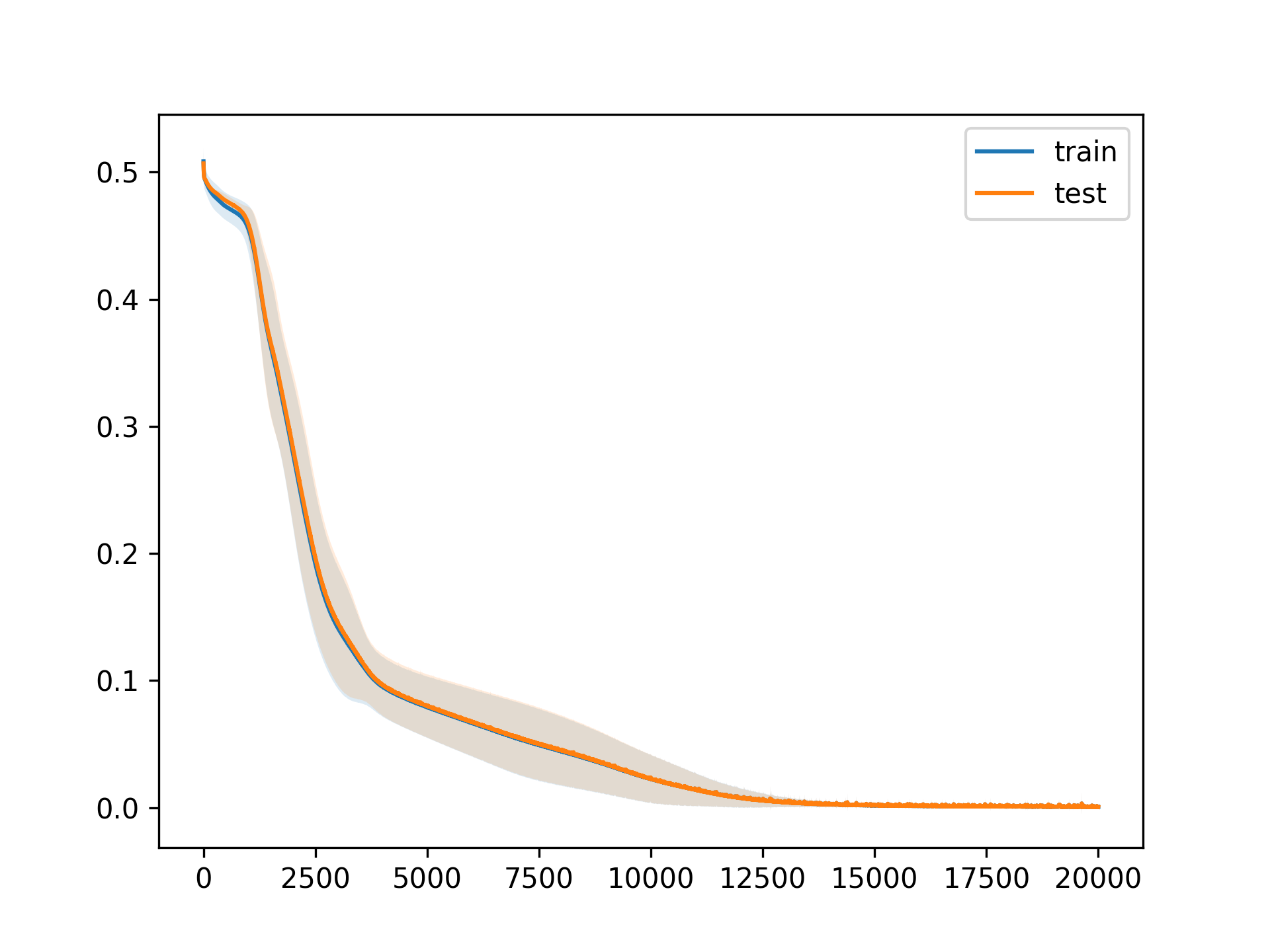}}
    \caption{(a) (b) is the train and test loss of $n=65$ and $n=1000$ nonuniform experiment, respectively. The activation function is $\mathrm{Tanh} x$.}
    \label{fig:one_dimension_loss_tanh}
\end{figure}

\begin{figure}[!h]
    \centering


    \subfloat{\includegraphics[width=\textwidth]{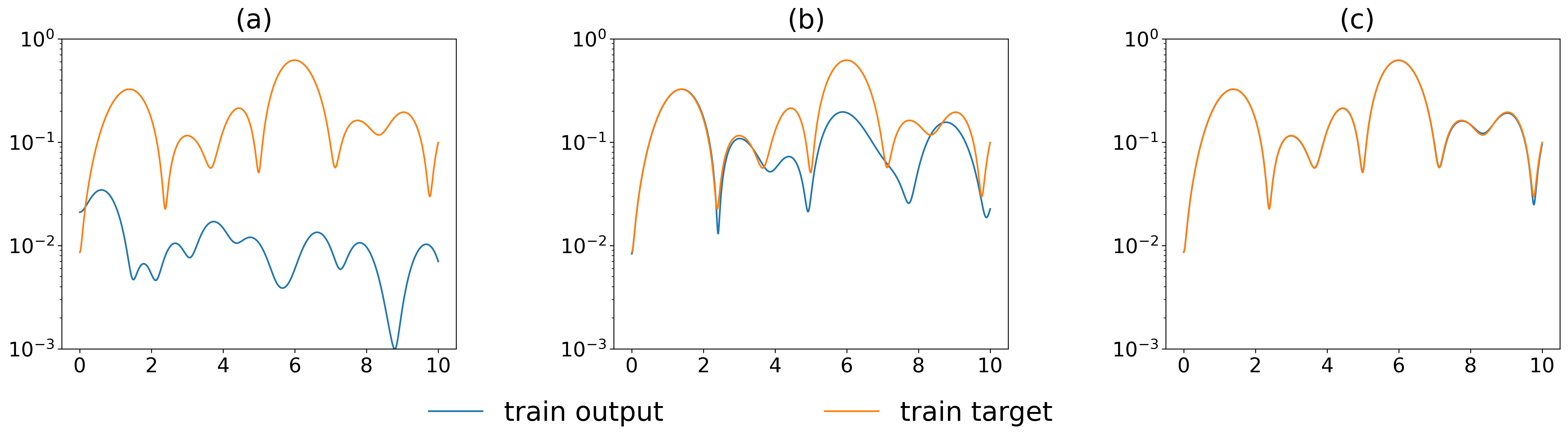}}\\
    \subfloat{\includegraphics[width=\textwidth]{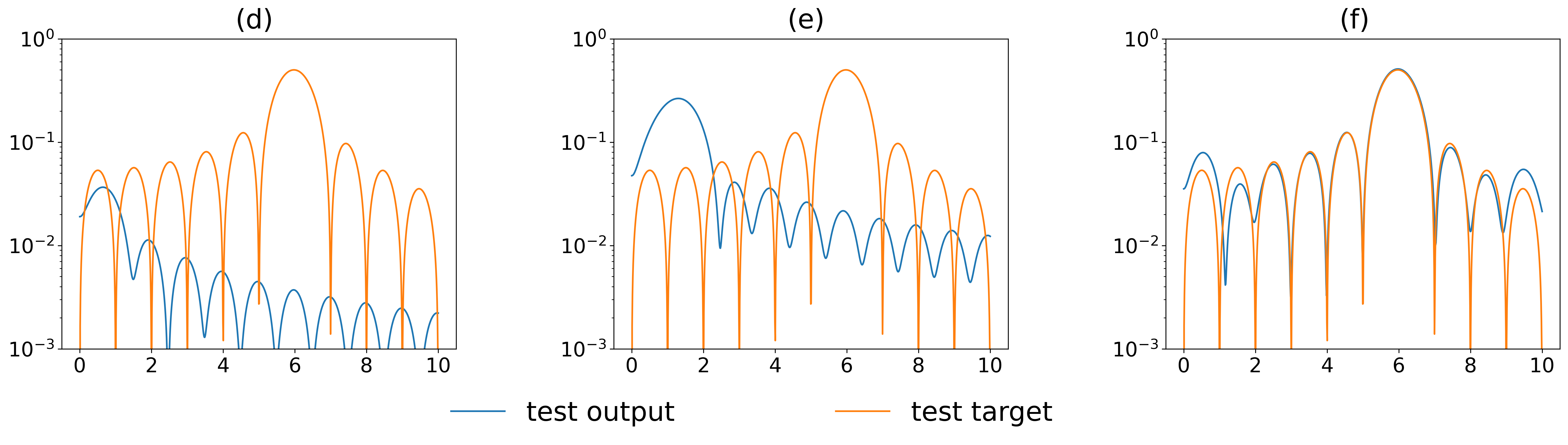}}\\
    \caption{The frequency spectrum during training of Fig.~\ref{fig:one_dimension_loss_tanh} (a). The columns from left to right correspond to epochs $0$, $1000$, and $18000$, respectively. The first row illustrates the frequency spectrum of the target function (orange solid line) and the network's output (blue solid line) on the training data. The second row shows the frequency spectrum of the target function (orange solid line) and the network's output (blue solid line) on the test data.}
    \label{fig:train_test_frequency_65_tanh}
\end{figure}

\begin{figure}[!h]
    \centering


    \subfloat{\includegraphics[width=\textwidth]{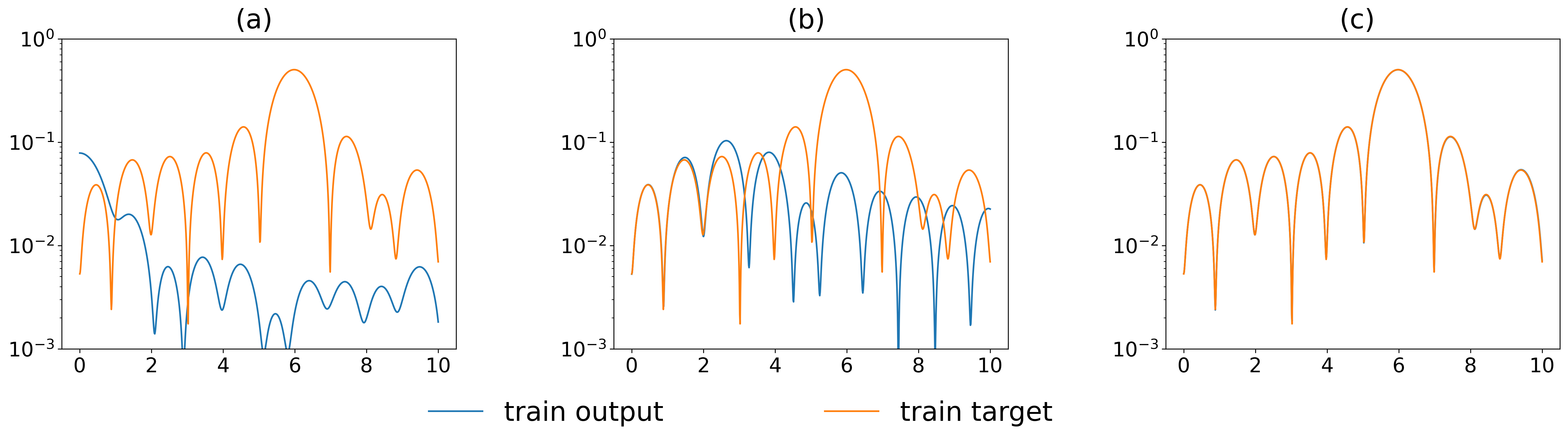}}\\
    \subfloat{\includegraphics[width=\textwidth]{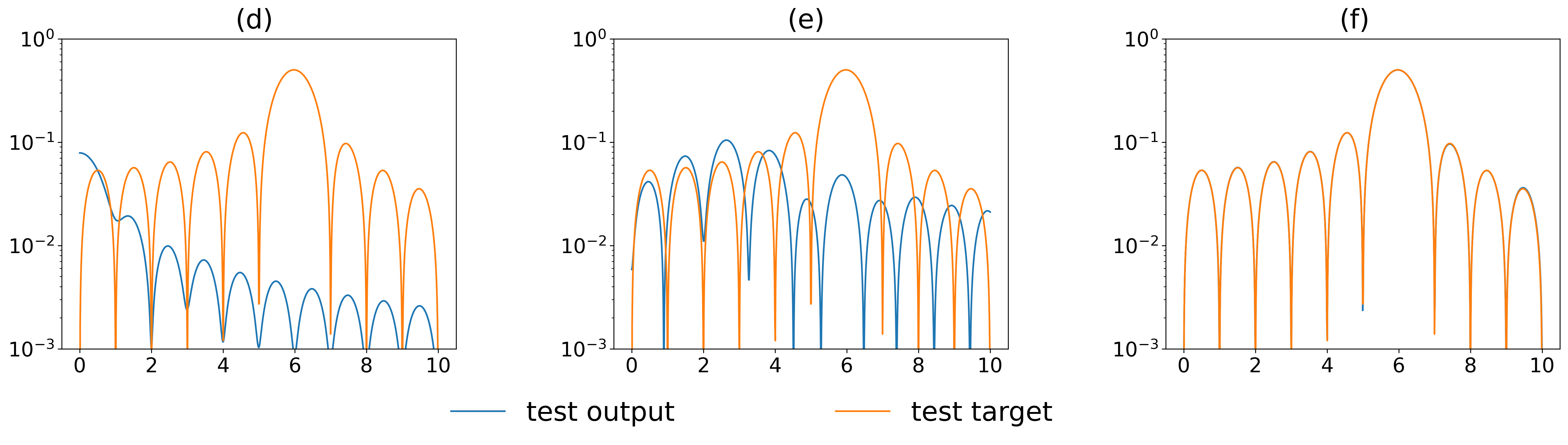}}\\
    \caption{The frequency spectrum during training of Fig.~\ref{fig:one_dimension_loss_tanh} (b). The columns from left to right correspond to epochs $0$, $1000$, and $18000$, respectively. The first row illustrates the frequency spectrum of the target function (orange solid line) and the network's output (blue solid line) on the training data. The second row shows the frequency spectrum of the target function (orange solid line) and the network's output (blue solid line) on the test data.}
    \label{fig:train_test_frequency_1000_tanh}
\end{figure}

Fig.~\ref{fig:one_dimension_loss_tanh} is the loss for $n=65$ and $n=1000$ nonuniform experiments when the activation function is $\mathrm{Tanh} $. Fig.\ref{fig:train_test_frequency_65_tanh} is the frequency spectrum of $n=65$ nonuniform experiment, and Fig.\ref{fig:train_test_frequency_1000_tanh} is the frequency spectrum of $n=1000$ nonuniform experiment.

We show that the mechanism of grokking in the frequency domain is the same varying in different activations.

\clearpage
\section{Errors in the parity function task}

The error function computes the difference between the label $y_i$ and $\mathrm{sign}(f(\mathbf{x}_i; \boldsymbol{\theta}))$ under MSE loss, indicating whether the neural networks have learned the parity function problem.

\begin{figure}[htb]
    \centering
    \subfloat[Error for Fig.~\ref{fig:parity_loss} (a) ]{\includegraphics[width=0.5\textwidth]{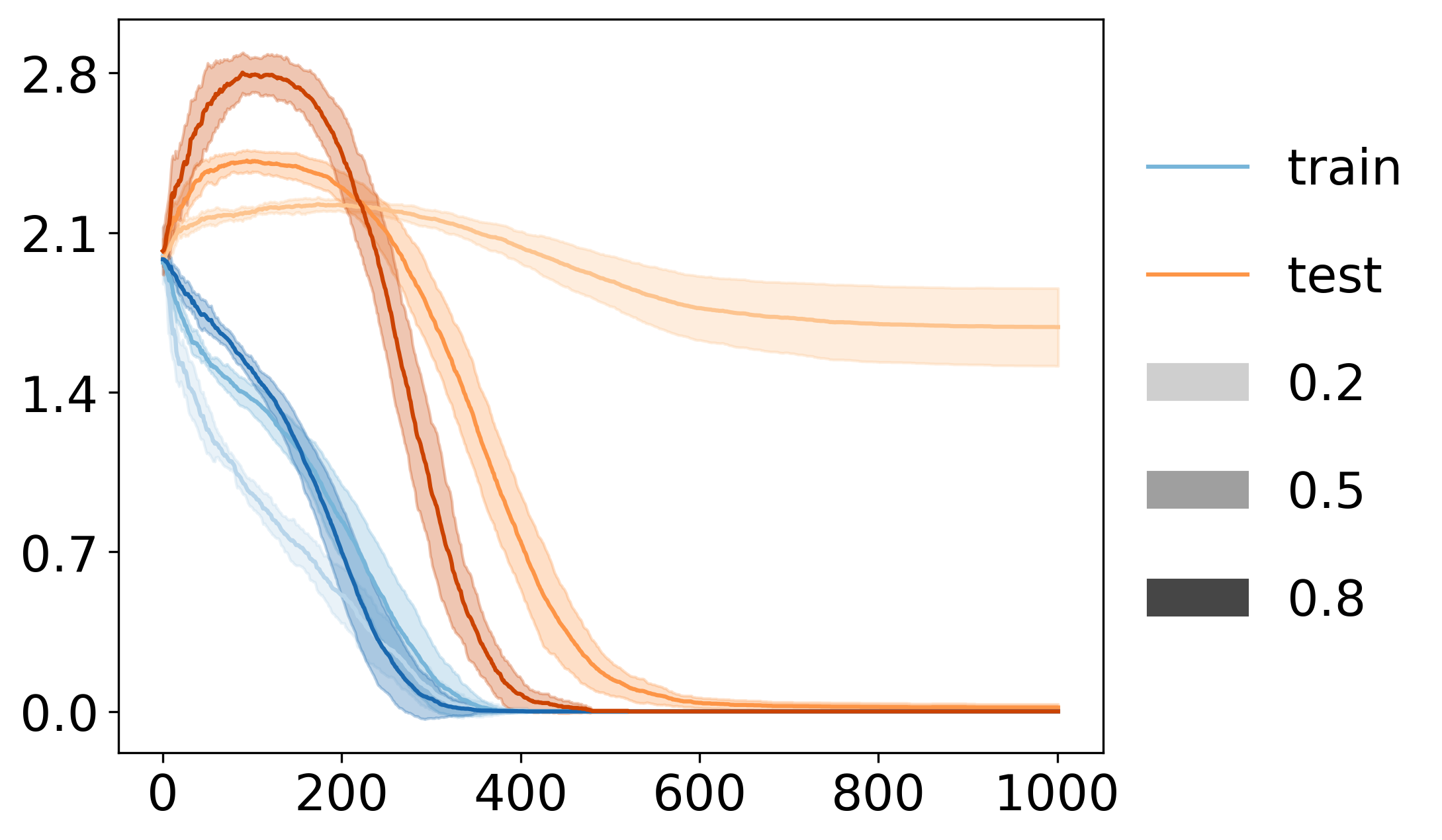}}
    \subfloat[Error for Fig.~\ref{fig:parity_process_test} (a)]{\includegraphics[width=0.5\textwidth]{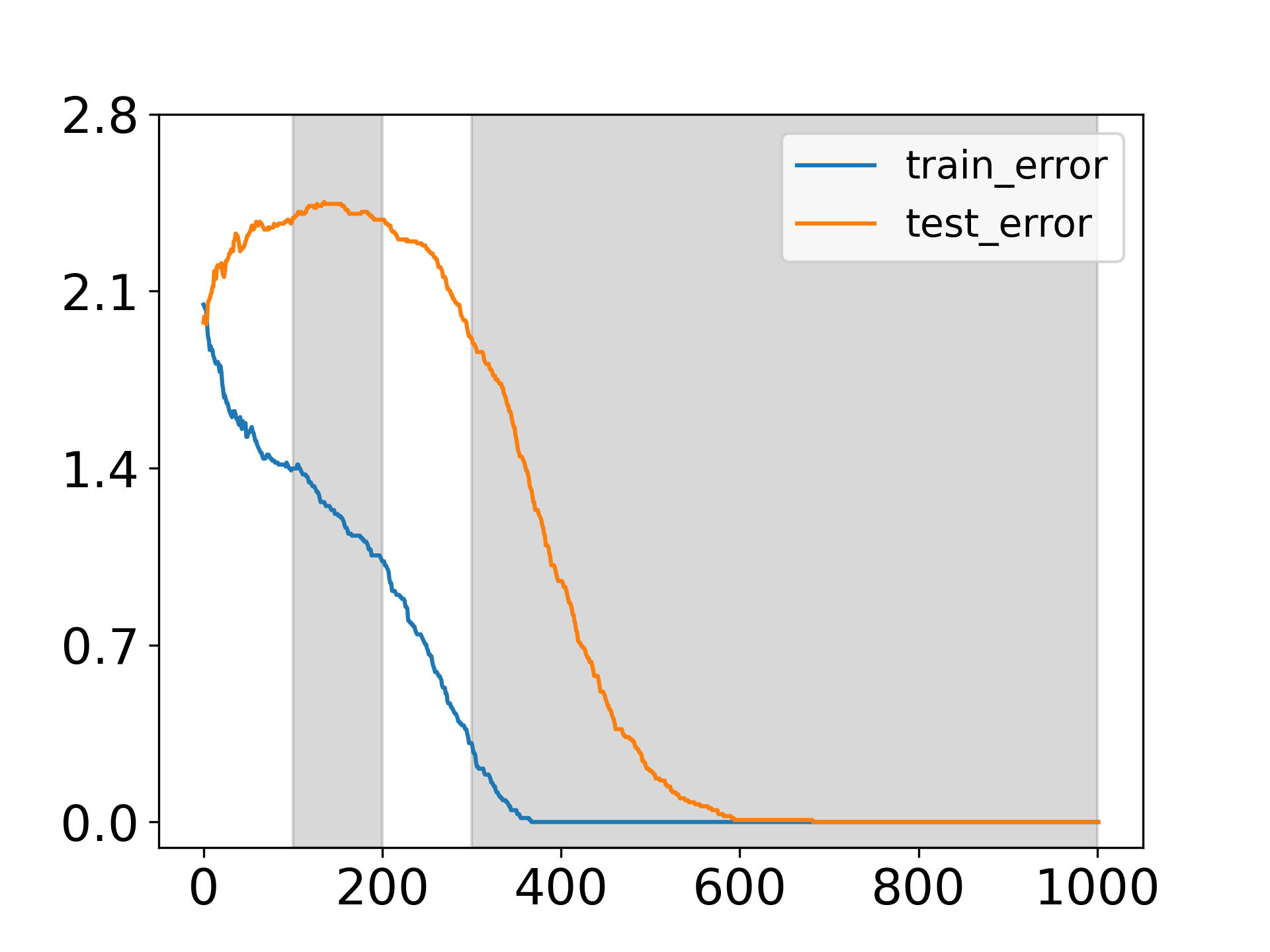}}
    \caption{The error for the experiments}
    \label{fig:parityfunction_error}
\end{figure}

\end{document}